\documentclass[letterpaper]{article} 
\usepackage[]{aaai25}  
\usepackage{times}  
\usepackage{helvet}  
\usepackage{courier}  
\usepackage[hyphens]{url}  
\usepackage{graphicx} 
\usepackage{natbib}  
\usepackage{caption} 
\usepackage{algorithm}
\usepackage{algorithmic}

\usepackage{subcaption}
\usepackage{placeins}
\usepackage{listings}
\usepackage{xcolor}
\usepackage{multirow}
\usepackage[inline]{enumitem}
\usepackage{amsmath}
\usepackage{booktabs}

\usepackage{soul}
\soulregister\cite7
\soulregister\ref7
\soulregister\pageref7
\usepackage{newfloat}
\usepackage{listings}
\DeclareCaptionStyle{ruled}{labelfont=normalfont,labelsep=colon,strut=off} 
\floatstyle{ruled}
\newfloat{listing}{tb}{lst}{}
\floatname{listing}{Listing}
\usepackage{xcolor}

\definecolor{highlighter}{RGB}{255,255,255}
\sethlcolor{highlighter}

\title{Clustering Internet Memes Across Similarity Dimensions}
\title{Clustering Internet Memes Through Template Matching and Multi-Dimensional Similarity}

\author{Hello}

\author{
 Tygo Bloem, Filip Ilievski
}
\affiliations{
    Department of Computer Science, Vrije Universiteit, Amsterdam, The Netherlands\\
    tygobloem@gmail.com, f.ilievski@vu.nl
}

\begin{document}

\maketitle

\begin{abstract}
Meme clustering is critical for toxicity detection, virality modeling, and typing, but it has received little attention in previous research. Clustering similar Internet memes is challenging due to their multimodality, cultural context, and adaptability. Existing approaches rely on databases, overlook semantics, and struggle to handle diverse dimensions of similarity. This paper introduces a novel method that uses template-based matching with multi-dimensional similarity features, thus eliminating the need for predefined databases and supporting adaptive matching. Memes are clustered using local and global features across similarity categories such as form, visual content, text, and identity. Our combined approach outperforms existing clustering methods, producing more consistent and coherent clusters, while similarity-based feature sets enable adaptability and align with human intuition. We make all supporting code publicly available to support subsequent research.
\end{abstract}

%
\begin{links}
    \link{Code}{https://github.com/tygobl/meme-clustering}
\end{links}

\section{Introduction}
Once niche in communities such as 4chan and Reddit, Internet memes have gained widespread popularity due to social media and advances in image editing software \cite{theisen.et.al.2020}. 
Memes condense complex ideas into shareable formats, typically using text overlayed on images \cite{Onielfa.et.al.2022}, reflecting cultural and social trends \cite{shifman2019internet}. They express humor and irony, but can also serve malicious purposes, such as coordinated hate campaigns and political messaging \cite{Pramanick.et.al.2021,qu.et.al.2023,beskow.et.al.2020}.

\begin{figure}[!t]
    \centering
    \begin{subfigure}[b]{0.3\textwidth}
        \centering
        \includegraphics[width=\textwidth]{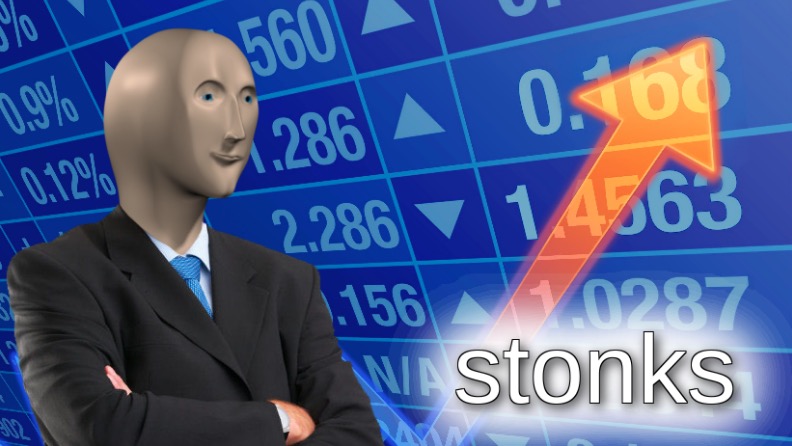}
        \caption{Original}
        \label{fig:example_original}
    \end{subfigure}
    \hfill
    \begin{subfigure}[b]{0.2\textwidth}
        \centering
        \includegraphics[width=\textwidth]{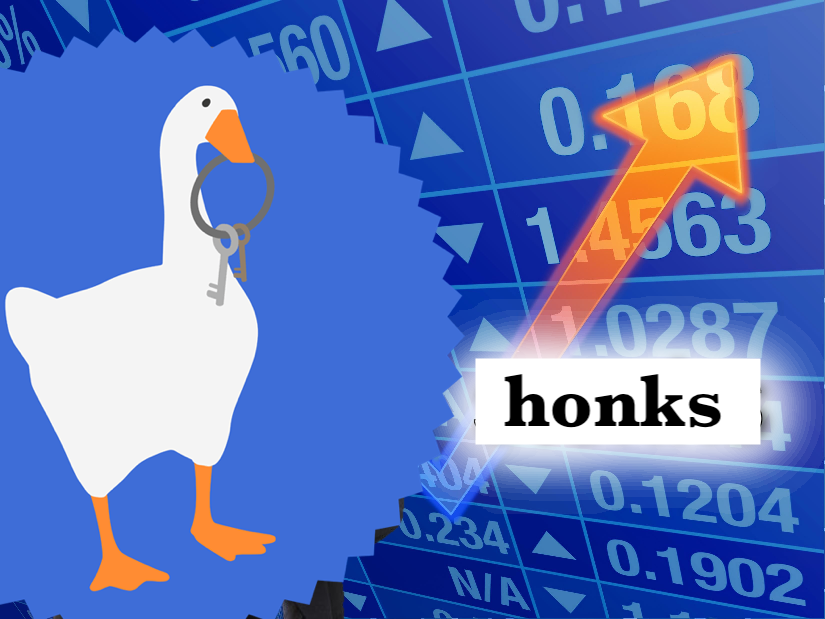}
        \caption{Form}
        \label{fig:example_form}
    \end{subfigure}
    \hfill
    \begin{subfigure}[b]{0.2\textwidth}
        \centering
        \includegraphics[width=\textwidth]{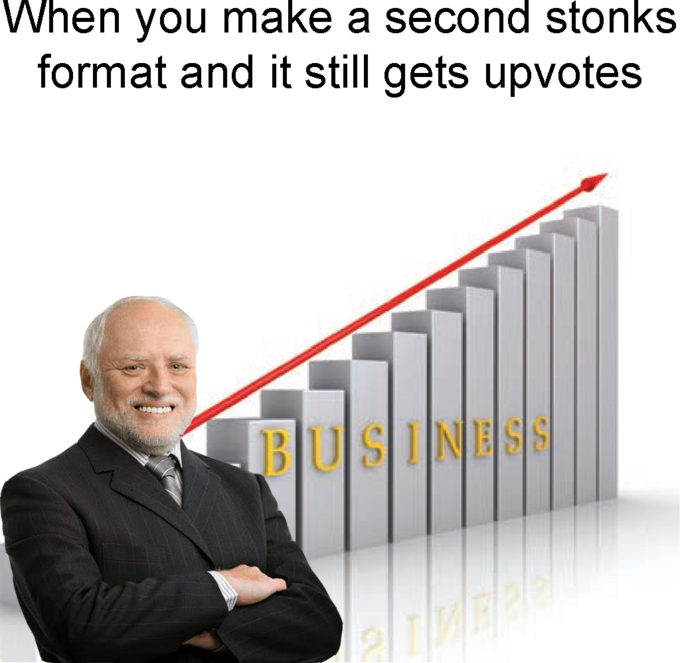}
        \caption{Visual Content}
        \label{fig:example_vcontent}
    \end{subfigure}
    \hfill
    \begin{subfigure}[b]{0.2\textwidth}
        \centering
        \includegraphics[width=\textwidth]{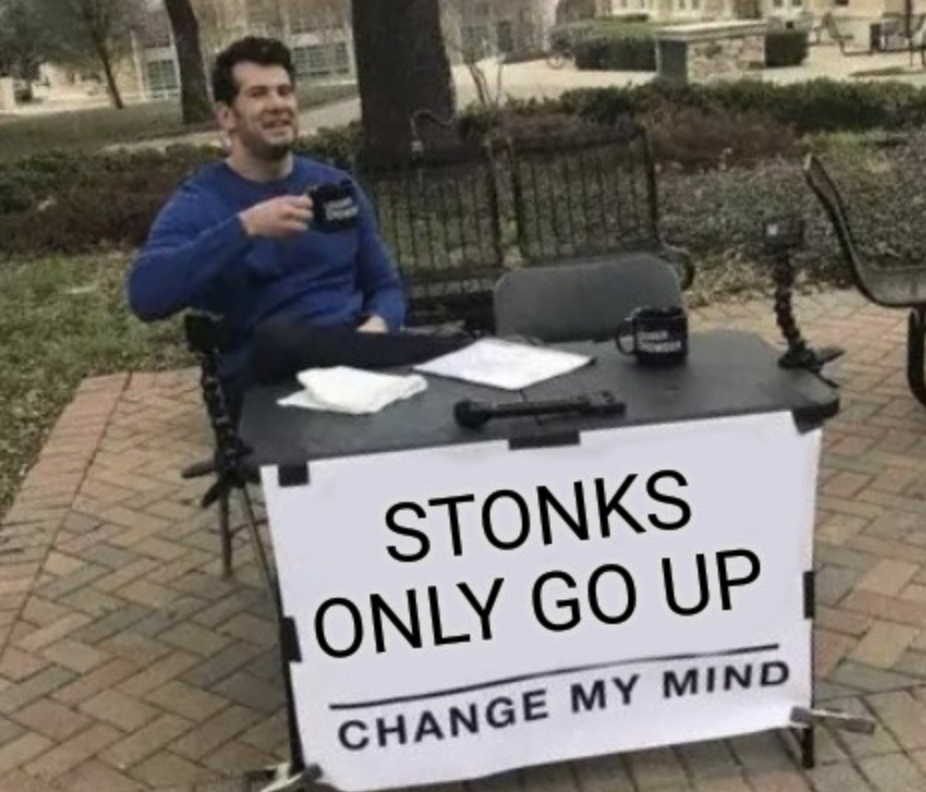}
        \caption{Textual Content}
        \label{fig:example_tcontent}
    \end{subfigure}
    \hfill
    \begin{subfigure}[b]{0.2\textwidth}
        \centering
        \includegraphics[width=\textwidth]{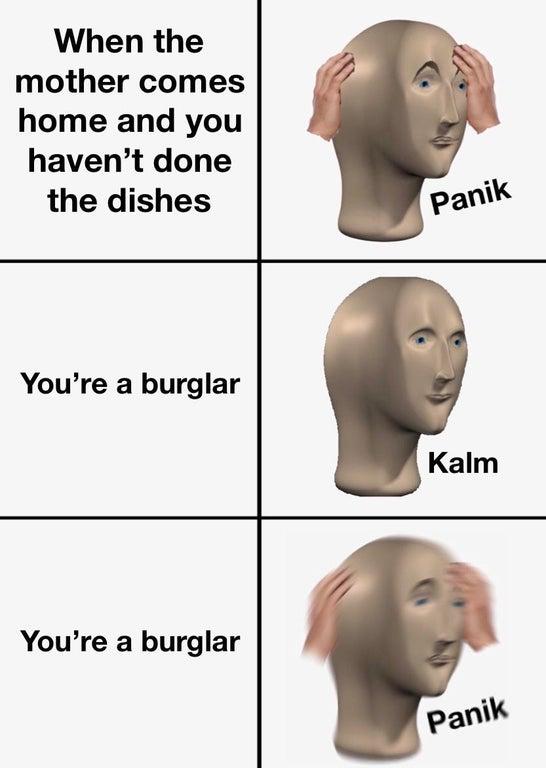}
        \caption{Identity}
        \label{fig:example_identity}
    \end{subfigure}
    \hfill
    \caption{The original ``Stonks'' meme template, accompanied by example memes related through various dimensions: form, visual and textual content, and identity.}
    \label{fig:examples_stonks}
\end{figure}

The rapid proliferation and diversity of memes online necessitate automated analysis. Yet, their subtle, multimodal nature \textemdash combining text, images, background knowledge, and context \textemdash \hl{requires a broad set of intelligence capabilities, making automated analysis of memes} an \textit{AI-complete challenge}~\citep{groppe2024way}. Research on Internet memes deals with modeling how memes evolve and spread in online communities \cite{Zannettou.et.al.2018,beskow.et.al.2020,qu.et.al.2023,Weng.et.al.2014}, analyzing the layered semantics and cultural contexts of memes (\citealp{Dancygier.Vandelanotte.2017,hee.et.al.2023}; \hl{\mbox{\citealp{liu-etal-2022-figmemes}}}), applying machine learning to downstream learning tasks such as identifying hateful memes \cite{kiela2021hateful,thakur2023multimodal,Grasso-et-al-2024}, and developing methods to generate novel memes \cite{perison.tolunay.2018,wang-wen-2015-cheezburger}. 
Analytical tasks with memes require a robust estimate of meme similarity and cluster discovery \cite{dubey.et.al.2018,theisen.et.al.2020}. \hl{Clustering memes can provide insights into how ideas and humor evolve online, aid moderation of harmful content, and enable detection of emerging trends, offering a valuable perspective on socio-political dynamics. Furthermore, given their challenging nature, memes can serve as an ideal testbed for developing context-sensitive methods broadly applicable to multimodal content analysis.}

Standard bottom-up clustering methods~\cite{qu.et.al.2023,Zannettou.et.al.2018} are based on pixel-level similarity captured through either local or global features, which offer complementary strengths~\cite{theisen.2022}. Recognizing that such methods \textit{cannot capture the multi-dimensional and rich semantics of memes}, another line of work grounds new memes in a knowledge repository of novel templates~\cite{joshi.et.al.2023,bates-et-al-2023}. However, existing template-based approaches \textit{rely on two strong assumptions: comprehensive database coverage and a single, form-driven similarity}.

A more fundamental challenge with all existing methods is the lack of agreement on how memes should be grouped, i.e., what it means for two memes to be similar.
Real-world memes \textit{defy strict template boundaries and exhibit partial similarity}~\cite{pandiani2024toxic}. 
New variants continually emerge, borrowing elements from other memes, mimicking, remixing by superimposing new elements, or even repurposing images entirely, which makes their clustering relative to the dimension of interest.
Figure \ref{fig:examples_stonks} illustrates the ``Stonks'' meme (a) alongside related memes:\footnote{\url{https://knowyourmeme.com/memes/stonks}} a remix of the original template, keeping only its \textbf{form} with new superimposed visual and text elements (Figure \ref{fig:example_form}); a mimetic variation, retaining the \textbf{visual content} of a person in suit observing rising stock values but substituting the subject with another well-known meme personality (Figure \ref{fig:example_vcontent}); a variation of a different meme template\footnote{\url{https://knowyourmeme.com/memes/steven-crowders-change-my-mind-campus-sign}} related through \textbf{text} by appropriating the ``Stonks'' catchword; and an \textbf{identity} similarity case \hl{(Figure \ref{fig:example_identity})} that features the same fictional \hl{character} (``Meme Man") \hl{as the ``Stonks'' meme} in an entirely different meme. These examples highlight the inherent challenges of matching these memes and defining what makes a coherent cluster. For example, Figure \ref{fig:example_tcontent} could be grouped with the ``Stonks'' meme, other memes that follow the Steven Crowder campus sign format, or both. Figure \ref{fig:example_identity} raises the question whether character continuity alone constitutes sufficient grounds for grouping. As prior work did not consider clustering relative to the similarity dimensions of interest, we note a gap between assumptions in the literature and real-world memes.
In summary, we note three challenges with prior work: the lack of consideration of semantics (bottom-up clustering methods), the reliance on databases (template-based methods), and the lack of customization for similarity dimensions (all methods).

\textit{How can we develop a comprehensive and modular meme clustering method that can natively adapt to similarity dimensions (without a knowledge base of templates)?} 
{\noindent Our \textbf{contributions} are twofold:
\begin{enumerate*}
    \item \textbf{A novel two-step approach} that unifies template-based identification and data-driven clustering. \hl{By first clustering a highly similar subset of memes to discover coherent templates, then matching additional memes to these templates, we achieve more accurate results than standard clustering methods without needing an external database}.
    \item \textbf{A modular framework that captures multiple dimensions of similarity} by consolidating both local and global features. \hl{We show how isolating each dimension (form, visual content, text, and identity) can be insightful, and how integrating them leads to robust clustering. This decomposition theoretically respects the intrinsically multimodal nature of memes, while also supporting applications such as dynamic meme retrieval.}
\end{enumerate*}



\section{Related Work}

Previous work on Internet meme clustering contributes new global and local features and grounds novel memes to known templates.

\noindent \textbf{Feature extraction for memes} is essential for typical 
data-driven, bottom-up approaches.
The scope of the features can be \textit{global} (encoding the overall image) or \textit{local} (capturing image regions) \cite{theisen.et.al.2020}. 
\citet{Zannettou.et.al.2018} design the \textbf{global} feature of perceptual hashing (PHASH). While PHASH effectively identifies near-identical images, it struggles with other semantically related but more visually distinct memes.
Convolutional neural network (CNN) models, such as MobileNet~\cite{howard2019searching} and VGG~\cite{simonyan2014very}, can also be used as global encoders of memes~\cite{theisen.2022}. 
Pre-trained multimodal encoders, like Contrastive language-image pre-training (CLIP), can learn
semantically rich and transferable global meme embeddings.
By fine-tuning CLIP on a dataset of memes from 4chan's /pol/ board, \citet{qu.et.al.2023} demonstrate its ability to identify hateful meme variants and potential influencers without explicit labels. Although CLIP can capture similarity beyond form without external databases, its noise levels in clustering remain high (47.9\%-62.2\%). 
\citet{theisen.et.al.2020} observe that relying on global features 
causes almost all images in the dataset to be grouped into a single cluster and is inadequate for remixed memes. 
They adapt the Speeded-Up Robust Features (SURF) algorithm, which identifies \textbf{local} points of interest and encodes them as rotationally invariant descriptor vectors.
The local features are indexed and used to construct an adjacency matrix and compute similarity based on the number of shared features.
While the results show that local features obtain superior coherence and interoperability over global features, they alone cannot capture the relations between the identified objects and the image form. 
To address these challenges, \citet{theisen.2022} integrate global (e.g., PHASH, MobileNet) and local (e.g., SURF) features, reporting
superior performance. \hl{Meanwhile, the method by \mbox{\citet{zhou-etal-2024-social}} clusters memes into templates by visually decomposing them, followed by semantic embedding of the meme text.}

Clustering with local and global bottom-up feature extractors captures low-level visual similarities (e.g., dogs or Facebook screenshots) and fails to preserve the core semantics of memes. In response, we extend the method to \cite{theisen.2022} \hl{with a discovery of} fundamental meme templates\hl{, while going beyond \cite{zhou-etal-2024-social} by utilizing a consolidated set of} popular features from the literature. Moreover, \hl{a key novelty of our work is our categorization} of the features in sets, which we leverage to \hl{modularize and adapt} meme matching \hl{according} to similarity dimensions.

\noindent \textbf{Template matching} relies on pre-existing repositories of known templates to identify and classify memes. 
\citet{Zannettou.et.al.2018} incorporate semantic information from KnowYourMeme (KYM) using keyword analysis to supplement the PHASH encoding. The study reports that focusing on near-identical images overlooks novel or evolving memes, evidenced by the high noise levels (62.8\%) and identified clusters that cannot be linked to KYM entries (15-24\%).
\citet{bates-et-al-2023} leverage meme templates from KYM to provide contextual grounding in classification. Following case-based reasoning, they assign the label of the most similar KYM template to each test meme.
\citet{dubey.et.al.2018} develop the MemeSequencer method for analyzing memes by decoupling overlaid information from identified template images. MemeSequencer extracts visual and textual features from each layer, capturing both the template's global context and the overlaid content's context.
\citet{Tommasinietal2023} developed the Internet Meme Knowledge Graph (IMKG), a knowledge base that enriches KYM with world knowledge and millions of meme instances to support complex queries about memes.
\citet{joshi.et.al.2023} propose to contextualize `memes in the wild' by grounding them to encyclopedic data and related entities in IMKG. Their approach matches novel memes with known templates using a Vision Transformer (ViT) \cite{Dosovitskiy.et.al.2020}. 

While data-driven methods fail to capture high-level semantics, template matching relies on external, manually curated knowledge repositories, which are incomplete and cannot capture the constantly evolving meme variations. A further challenge with all presented methods is their focus on form-based similarity without an adaptive mechanism to address the multifaceted relationships between memes. Inspired by these observations, we introduce a method that combines the strengths of both template-based matching and feature-based approaches. We define a comprehensive set of feature extraction techniques that capture diverse similarity dimensions and automatically identify meme templates from the data without external knowledge repositories.

\begin{figure*}
    \centering
    \includegraphics[width=\linewidth]{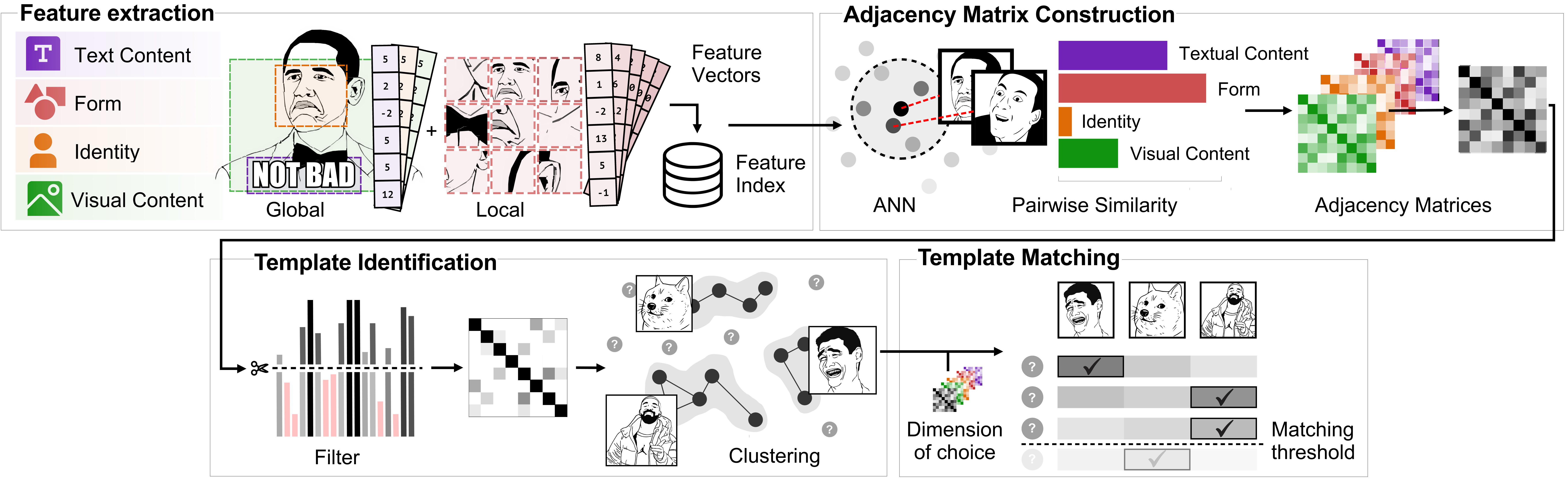}
    \caption{Our methodology: feature extraction, adjacency matrix construction, template identification, and template matching.}
    \label{fig:method}
\end{figure*}

\section{Methodology}

Our methodology comprises four steps (Figure \ref{fig:method}). The first phase produces feature vectors that serve as input to create multiple adjacency matrices aligned with meme dimensions. The adjacency matrices are leveraged to identify templates in a bottom-up manner, ultimately used to cluster a meme in the template matching step.

\subsection{Feature Extraction}

We consolidate techniques for extracting global and local image features as vector embeddings. Global features represent entire images, and local features encode key regions or points. Inspired by \hl{previous} work \cite{theisen.2022,thakur2023multimodal}, we use multiple feature extractors to capture complementary semantics of form, content, and identity. 

\noindent \textbf{Global features.}
We use four global feature extractors \hl{that focus} on high-level semantics, overall structure and content, color distribution, and text.  
\begin{enumerate*}
    \item To extract high-level semantic features such as objects (e.g., cat, car), scenes (e.g., cityscape, beach), abstract concepts (e.g., happiness, danger), and their interactions, we employ \textbf{ViT} \cite{Dosovitskiy.et.al.2020}.  \hl{We deliberately choose a base ViT rather than a text-image model such as CLIP \cite{qu.et.al.2023}, because we aim to keep the purely visual semantics separate from the textual overlays. Preliminary experiments revealed that CLIP may overemphasize superimposed text in images, merging textual and visual dimensions.}
    \item To capture the overall structure and content of an image, we leverage \textbf{Perceptual Hashing (PHASH)} \cite{klinger2013phash}: a compact, binary fingerprint of an image that represents its visual features. These features are particularly effective in detecting meme templates with slight text variations or edits \cite{Zannettou.et.al.2018}. 
    \item To capture color density, we compute \textbf{color histograms}. They capture image color density in the hue, saturation, and value (HSV) color space, enabling similarity calculation of images based on their color distribution regardless of spatial arrangement. Memes with largely overlapping color profiles may indicate that they share a common base template image \cite{morinan-2021}. Color histograms complement PHASH in targeting the form of memes, as the latter does not account for color and is not as robust to rotation and cropping. 
    \item To encode text content within memes, we use \textbf{BERT} \cite{devlin2019}. We assign greater weight to specific parts of the textual content to prioritize memes with familiar catchphrases and phrasal templates. We identify frequent bigrams in the meme dataset's Optical Character Recognition (OCR) text and double their word weights.
\end{enumerate*}

\noindent \textbf{Local features.}
We use two extractors of local features.
\begin{enumerate*}
    \item To extract facial features, we use a pre-trained \textbf{face recognition model} from the Dlib library, which is optimized for speed and accuracy \cite{dlib}. We use facial landmark features to relate memes based on the identity of the people or fictional characters they feature.
    \item We employ \textbf{SURF} \cite{bay.et.al.2006} to identify \hl{points of interest in the} images and construct a 64-dimensional descriptor vector for each. These descriptors are rotationally invariant and robust to changes in illumination, making them ideal for meme analysis, where small, distinctive elements (e.g., specific characters, gestures, or logos) are critical to linking remixed memes together. To avoid overemphasizing letters and fonts, we superimpose black boxes over text regions before extracting the features (see the Appendix for further details). 
\end{enumerate*}

\subsection{Adjacency Matrix Construction}

\noindent \textbf{Feature-specific matrices.}
We encode each meme in an index using its features, which allows for an approximate nearest neighbor (ANN) search based on cosine similarity.  
The distances to the neighbors in the index are used to calculate similarity scores and populate the corresponding cells in the adjacency matrix. We use a hyperbolic tangent to convert a distance $d$ to a similarity score: $s(d) = 1 - \tanh(d) $. The function is bounded between 0 and 1 and rewards smaller distances. The similarities are used to construct a graph representation using an adjacency matrix $A$ for each feature set, following \cite{theisen.2022}. $A$ is an $N \times N$ matrix,  where \( N \) is the total number of memes, and each cell \( A_{i,j} \) represents the similarity between memes $i$ and $j$. \hl{We set to zero any similarity below a threshold (set to 0.001) to reduce noise and maintain computational efficiency, resulting in a sparse adjacency matrix that focuses on meaningful relationships.} \hl{We apply L2 normalization to each feature vector before computing similarities, enabling consistent similarity computations and meaningful comparisons across feature spaces.} 

\noindent \textit{Global Features.} A single global feature vector represents each image. Let \( \mathbf{g}_i \) be the global feature vector for the image \( i \). We perform an ANN search in \( \mathcal{I}_{\text{global}} \) using the global feature \( \mathbf{g}_i \) to find the nearest global features. This search results in a set \( \mathcal{R}_i \) containing the nearest global features to \( \mathbf{g}_i \). Each element in \( \mathcal{R}_i \) is a global feature \( \mathbf{g}_j \) corresponding directly to another image \( j \). We convert the distances resulting from the ANN search into similarity scores, \( s(\mathbf{g}_i, \mathbf{g}_j) \), where \( \mathbf{g}_j \in \mathcal{R}_i \). Formally: \[ A[i, j] = s(\mathbf{g}_i, \mathbf{g}_j) \]

\noindent \textit{Local Features.} Let \( \mathbf{f}_{i,k} \) denote the \( k \)-th local feature vector of image \( i \). For each local feature, we perform ANN search \( \mathcal{I}_{\text{local}} \) to find a set of nearest features, denoted as \( \mathcal{R}_{i,k} \).  Each element in \( \mathcal{R}_{i,k} \) corresponds to a feature vector \( \mathbf{f}_{j,l} \) from potentially any image \( j \) and any feature index \( l \). We convert the distances resulting from the ANN search into similarity scores, \( s(\mathbf{f}_{i,k}, \mathbf{f}_{j,l}) \), where \( \mathbf{f}_{j,l} \in \mathcal{R}_{i,k} \). For each retrieved feature \( \mathbf{f}_{j,l} \), we identify the image \( j \) it belongs to, and add the similarity score \( s(\mathbf{f}_{i,k}, \mathbf{f}_{j,l}) \) to the adjacency matrix entry \( A[i, j] \). Hence, the entire process can be summarized by the following formula for updating the adjacency matrix:
\[ A[i, j] = \sum_{k=1}^{M} \sum_{\mathbf{f}_{j,l} \in \mathcal{R}_{i,k}} s(\mathbf{f}_{i,k}, \mathbf{f}_{j,l}) \]

\noindent \textbf{Matrix aggregation.}
Creating separate adjacency matrices for each feature set enables flexible analysis of image similarities based on individual or combined features. \hl{Each adjacency matrix is constructed by computing similarity scores between normalized feature vectors, ensuring a balanced contribution from different features.} Since each cell in these matrices corresponds to the same meme pair, \hl{we can aggregate them by summing their values element-wise.} This modular approach allows us to target specific similarity dimensions regardless of their granularity, namely:
\begin{enumerate*}
    \item \textbf{Form} connects memes that share \hl{surface-level visual attributes} such as images, colors, fonts, and design elements. We sum the adjacency matrices of PHASH, Color Histograms, and SURF features for a comprehensive form similarity measure. \hl{It doesn’t necessarily reflect what specific objects or people appear in the image, but rather focuses on aesthetic features.}
    \item \textbf{Visual content} \hl{identifies deeper visual similarities by considering the semantic content depicted in the memes.} Specifically, it captures if memes share recognizable objects, similar objects, facial expressions, \hl{actions, or environments.} We use the adjacency matrix with ViT features for this dimension.
    \item \textbf{Textual content} connects memes with similar captions using the adjacency matrix with BERT features.
    \item \textbf{Identity} links memes featuring the same real-world individuals or fictional characters using the adjacency matrix constructed from the Face Landmark features.
\end{enumerate*}
While we recognize that this four-way classification of similarity may be disputable, it still provides a valuable lens to distinguish between dimensions of similarity.
Finally, we construct a \textbf{combined} adjacency matrix that combines all individual matrices, offering a comprehensive similarity measure across all dimensions and potentially highlighting high similarity within a single dimension. \hl{We directly sum the feature-specific matrices to keep the aggregation simple and in light of the effectiveness of this approach in our experiments. We leave experiments with weighted averaging for future work.}

\hl{Our selection of similarity dimensions builds upon established distinctions and empirical observations in the literature. Separating \textit{text} content from visual elements is fundamental in multimodal analysis, including research on memes \mbox{\cite{dubey.et.al.2018}}. Within the visual domain, previous work employs features capturing either low-level structural or pixel similarity like PHASH and SURF, as seen in \mbox{\cite {Zannettou.et.al.2018,theisen.et.al.2020}}, which we term \textit{form}; or higher-level semantic concepts like objects and scenes using deep models, for example, ViT and CNNs \mbox{\cite{joshi.et.al.2023,theisen.2022}}, corresponding to our \textit{visual content} dimension. We make this distinction explicit, recognizing form and content as different facets of visual similarity. Finally, we added \textit{identity} based on the empirical prevalence of memes centered on recurring individuals or characters, as well as a recent competition highlighting their importance for memes \mbox{\cite{Sharma_2022}}.}

\subsection{Template Identification} 
\noindent \textbf{Matrix filtering.}
To identify templates, we first cluster using a filtered adjacency matrix. \hl{By filtering out memes with low similarity scores, we obtain a smaller subset of images with robust edges, effectively bootstrapping a set of strongly coherent meme clusters to serve as base templates. Since these clusters are smaller but more semantically focused, they provide a strong foundation for subsequently matching the remaining images to them. As a result, we expect higher precision compared to an alternative approach where we simply cluster all images at once.} \hl{Filtering is always applied to the full aggregated adjacency matrix under evaluation.}

Specifically, we remove all pairwise similarities \hl{below} a fixed threshold, $\theta$. This filtering technique was selected to facilitate comparison between methods, such as assessing the marginal benefits of using the combined feature set versus only local features. However, it does not fully exploit the potential advantages of using a richer combined feature set, such as an alternative filtering technique that would retain only the connections between images with high pairwise similarities across more than one dimension. Such an approach could potentially produce \hl{a} more robust template identification. \hl{While filtering the adjacency matrix (using $\theta$) effectively identifies core templates, a potential weakness is overlooking subtle variations or nascent meme trends. Given the dynamic nature of memes and the rapid emergence of meme templates, we opt for an unsupervised rather than a supervised approach based on train-test splits to avoid overfitting. We leave it to future work to explore alternative matrix filtering techniques, including supervised methods.}

\noindent \textbf{Matrix-to-graph conversion.}
The filtered matrix is converted to a graph structure for Louvain clustering~\cite{Blondel_2008}, a hierarchical algorithm based on graph modularity. Louvain clustering optimizes the connection density within communities versus between communities by iteratively moving nodes between communities until a stable, high-modularity partition is achieved. 
Louvain clustering is adequate for sparse graphs and produces a more balanced distribution of meme images across clusters compared to competitors \hl{such as} Markov and Spectral clustering \cite{theisen.2022}.

\subsection{Template Matching} 
We match memes to the identified templates by calculating meme-template similarity, assigning memes to templates, and ranking trade-offs. This methodology allows for systematic and quantitative control over meme clustering, optimizing the balance between intra-cluster similarity and the breadth of images included in the clusters.

\noindent \textbf{Template vector calculation.}
Let \( T = \{T_1, T_2, \ldots, T_k\} \) represent the set of \( k \) template clusters. For each template cluster \( T_i \), we construct a similarity vector 
   $S_i = [s_{i1}, s_{i2}, \ldots, s_{im}]$.
Here, \( s_{ij} \) denotes the similarity score between the \( j \)-th image \( I_j \) and all members of the template cluster \( T_i \), computed as the average of the corresponding adjacency matrix cells. Formally, if \( T_i = \{T_{i1}, T_{i2}, \ldots, T_{in_i}\} \) and the adjacency matrix is \( A \), then:
   \[
   s_{ij} = \frac{1}{n_i} \sum_{k=1}^{n_i} A[j, T_{ik}]
   \]
where \( A[j, T_{ik}] \) is the similarity score \hl{of} the adjacency matrix between the image \( I_j \) and the \( k \)-th member of \hl{the} template \( T_i \).

\noindent \textbf{Meme assignment.}
For each image \( I_j \), we determine its position within each similarity vector \( S_i \) and select the template cluster \( T_i \) that yields the maximum similarity score:
   \[
   \text{max\_sim}_j = \max_{i} (s_{ij})
   \]
   Consequently, image \( I_j \) is assigned to the template cluster \( T_{i^*} \) where:
   \[
   i^* = \arg\max_{i} (s_{ij})
   \]

\noindent \textbf{Incremental ranking.}
Finally, after determining the maximum similarity score \( \text{max\_sim}_j \) for each meme \( I_j \), we rank all memes in descending order based on their maximum similarity scores:
   \[
   \text{rank}(I_j) = \text{sort\_desc}(\text{max\_sim}_j)
   \]
   where \( \text{sort\_desc} \) denotes the sorting operation in descending order. To incrementally cluster images, we proceed through the ranked list from highest to lowest \( \text{max\_sim}_j \). This process prioritizes the images with the highest coherence to any template cluster, thereby controlling the trade-off between cluster coherence and the number of memes clustered.

\section{Experimental Setup}

\subsection{Data}

Our evaluation is based on the union of \hl{KYM and Reddit data}, both popular sources for existing datasets~\cite{joshi.et.al.2023,Zannettou.et.al.2018}. As existing clustering datasets focus on size~\cite{Zannettou.et.al.2018} or assume a comprehensive knowledge base~\cite{joshi.et.al.2023}, we opt to create our dataset to control its size and density properties (memes per template).

\textbf{KYM} is a large crowdsourced repository of meme knowledge, offering example images, detailed descriptions, and metadata for thousands of memes.\footnote{\url{https://knowyourmeme.com}} For this study, we scraped the 150 most popular meme entries as of February 2024, extracting up to 100 static images per entry while excluding videos and GIFs. Each entry contains at least 10 example images, with an average of 73 images per entry, resulting in 10,917 images. Sourcing multiple example images per meme entry allows us to obtain groups of related memes in our relatively small dataset rather than many random, unrelated memes. This will help seed the clustering algorithms applied later in the study.
Furthermore, the images are already linked to specific meme entries, which can serve as a ground truth to validate the consistency of the clustering of our method and \hl{the} relevant baselines. 
Complementing the KYM data, we gathered 9,869 memes from the r/memes subreddit on \textbf{Reddit}, capturing memes `in the wild' as they circulate organically within online communities.\footnote{\url{https://www.reddit.com/r/memes/}} This dataset spans from 2011 to 2024, with a higher concentration of memes in recent years.  The memes were collected using a Reddit data dump from Pushshift.io.\footnote{\url{https://the-eye.eu/redarcs/}} Posts were randomly sampled, and the corresponding images were scraped. 

In total, our data collection yields 20,786 memes.
Collecting data from KYM and a dedicated meme community on Reddit ensures high confidence that the collected images are memes, eliminating the need for preprocessing steps to filter out non-meme content. We verified this for the Reddit dataset by manually inspecting a random sample of 100 images for memes. 96 of these images were memes; the other 4 were fan art and miscellaneous photos.

\subsection{Tasks and Metrics}

In line with our contributions, we assess two aspects of the clusters produced by our method. First, we investigate the accuracy of our method compared to ablated baselines by computing the cluster \textbf{consistency} against preexisting clusters in KYM and the cluster \textbf{coherence} through an Imposter-Host task with a human study. Then, we investigate the \textbf{effect} of various similarity-based feature sets and their \textbf{alignment} with human annotations of similarity dimensions.

\noindent \textbf{Consistency with existing clusters} 
assumes the existence of golden cluster labels, which are only available for the KYM portion of our data. 
Here, we exclude clusters with fewer than three KYM images.
We use KYM meme entries as a proxy for desirable clusters to determine the quality of the clustering methods. Specifically, we evaluate how many images within each cluster correspond to the same KYM meme entry. We \hl{measure} performance at predetermined intervals of 5,000, 8,500, and 11,000 clustered images.
To quantify how closely the system clusters align with KYM, we define the \textit{consistency} of a cluster \( C_t \) as follows:
\[
\text{Consistency}_{C_t} = \frac{\max\limits_{k} \left( n_{k,t} \right)}{\sum\limits_{j} n_{j,t}}
\]
where \( C_t \) denotes cluster \( t \) and \( n_{k,t} \) represents the number of images in cluster \( t \) that belong to the \( k \)-th template. We calculate the \textit{average consistency} by weighting clusters according to size for a fair comparison across different approaches.

\noindent \textbf{Cluster coherence}
relies on human signals to avoid data biases introduced by the KYM community. Following \cite{theisen.et.al.2020}, we conduct the Imposter-Host task. Here, we present five images to a human judge: four from the same ``host" cluster and one ``imposter" image randomly selected from another cluster. The human's objective is to identify the imposter image: assuming the images within a cluster are visually coherent, the outlier should be easily discernible. 
This evaluation involves 12 non-experts varying in age, gender, and background (see \hl{the} Appendix for further details). 
The clusters were randomly sampled to ensure comprehensive dataset coverage and minimize potential bias.
We evaluate the clustering methods across various increments of the total clustered images, rewarding them for agreeing \hl{with humans on the imposters}. This setup \hl{allows us to} observe performance trends by incorporating images with progressively lower similarity. The chosen intervals are the same as in the consistency evaluation and include both KYM and Reddit memes.
We use accuracy as a metric, as is common in multiple-choice QA tasks. Accuracy scores are weighted by the cluster size.

\noindent \textbf{Effect of similarity dimensions.}
In this task, the same 12 humans are presented with five memes, all belonging to the same cluster according to the output of a clustering method. Among these, one meme is the least likely member of the cluster, as \hl{it was} the last one added by our method. The human's objective is to answer ``yes" or ``no" to the question ``Are all the memes above related?" We choose memes \hl{in} random ranks $\text{rank}(I_j)$ between 0 and 5000. A moving average accuracy of image matching, weighted by cluster size, is computed across ranks. For each rank representing the cumulative number of images matched, the accuracy is calculated over a sliding window of 1500 ranks, with larger clusters contributing more to the accuracy measurement. 
We expect accuracy to gradually decrease for higher-ranked images because these matches are found later in the process and have lower similarity scores to the templates.

\noindent \textbf{Human alignment.} When \hl{a} human selects ``yes'', indicating that all memes appear related, they are randomly prompted 50\% of the time to specify how the memes are related: by form, visual content, text, or identity.
This step helps us \hl{to} determine \hl{whether} our features accurately represent the intended similarity dimension.
An example of the task interface and instructions \hl{is} provided in the Appendix. 

\hl{We opted for broadening the annotation coverage through randomization to evaluate performance across a wide variety of generated clusters with a practical number of annotators, rather than focusing intensely on a small, potentially unrepresentative subset. In this setup, each annotation task is unique, and no data point is annotated by multiple annotators, which hinders our ability to compute inter-annotator agreement indicators. Instead, we manually inspected a small sample of the annotations to confirm their validity informally, noting high agreement with some natural variation of how individuals perceive similarity. We anticipate a follow-up study specifically designed with overlapping annotations to calculate inter-annotator agreement to provide further insights into how people perceive meme similarity. }

\subsection{Baselines}

Consistency and coherence are compared between template-based and standard bottom-up clustering methods. \textit{Template-based clustering} is our proposed method that involves initially identifying templates through stringent clustering and matching memes to these templates incrementally. \textit{Standard clustering} directly applies the clustering algorithm (e.g., Louvain) to the adjacency matrices without identifying templates first, following \cite{theisen.2022}. The values in the adjacency matrix are filtered at different percentiles and re-clustered to test performance at increments of clustered images. 
While Louvain clustering fits our task well, to demonstrate the generalizability of our findings, we also experiment with another clustering algorithm, DBSCAN.
For both standard and our template-based clustering, and both Louvain and DBSCAN, we compare their combined feature sets against ablations using only the four global, the two local, or the ViT features to determine how clustering quality is affected when only one similarity dimension is covered. \hl{In addition, besides ViT, we also include CLIP as a baseline, which is a natural choice for meme encoding given its multimodal training.}

As the coherence experiment requires laborious human validation, we include a smaller set of baselines: the combined baseline with standard features and the best-performing template-based baseline.
The effect and human alignment of similarity dimensions between text, visual content, form, and identity features \hl{are analyzed}. We match memes with the templates identified for each similarity dimension based on \hl{their} corresponding feature set.

For a fair comparison between methods, we set $\theta$ so that the final number of images within the identified templates always totals 5,000, although the number of templates may vary. 
The total of 5,000 images was determined by manual inspection to reflect the point at which the templates still predominantly exhibit minor variations in superimposed text or small visual elements while maintaining the core image structure.

\section{Results}

We compare the consistency and coherence of the clusters produced by our method \hl{with} baselines.
Subsequently, we test the effect of our feature sets in matching memes to templates and their alignment with human similarity judgments. \hl{Finally, we present two case studies that show the broad applicability of our methodology.}

\subsection{Clustering Accuracy}

\begin{table}[!t]
\centering
\begin{tabular}{l l r r r}
\toprule
\textbf{Clustering} & \multirow{2}{*}{\textbf{Feature set}} & \multicolumn{3}{c}{\textbf{\# Images clustered}} \\ \cmidrule(l){3-5} 
\textbf{method} &  & \textbf{5000} & \textbf{8500} & \textbf{11000} \\ \midrule

\multirow{5}{*}{Standard} & \hl{CLIP} & \hl{0.51} & \hl{0.40} & \hl{0.31} \\
& ViT    & 0.71             & 0.48             & 0.35   \\
& Global   & 0.84              & 0.61             & 0.48              \\
& Local    & 0.84                 & 0.70             & 0.53              \\ 
& Combined & \textbf{0.94}               & 0.67             & 0.54               \\
\midrule
 & \hl{CLIP} & \hl{0.51} & \hl{0.66} & \hl{0.70} \\
Template & ViT & 0.71 &   0.68  &  0.65 \\
-based & Global  & 0.84 &   0.81  &  0.78              \\
(ours) & Local   & 0.84  &  0.84 &   0.79        \\
& Combined & \textbf{0.94}    & \textbf{0.89}    & \textbf{0.87}     \\

\bottomrule
\end{tabular}
\caption{Consistency scores across methods and \# images clustered when using \hl{the} Louvain clustering algorithm. 
The best results are shown in bold. The number of clusters for various feature sets\hl{, together with additional metrics and baselines,} is given in the Appendix.}
\label{tab:clustering}
\end{table}

\begin{figure}[!t]
    \centering
    \includegraphics[width=0.93\linewidth]{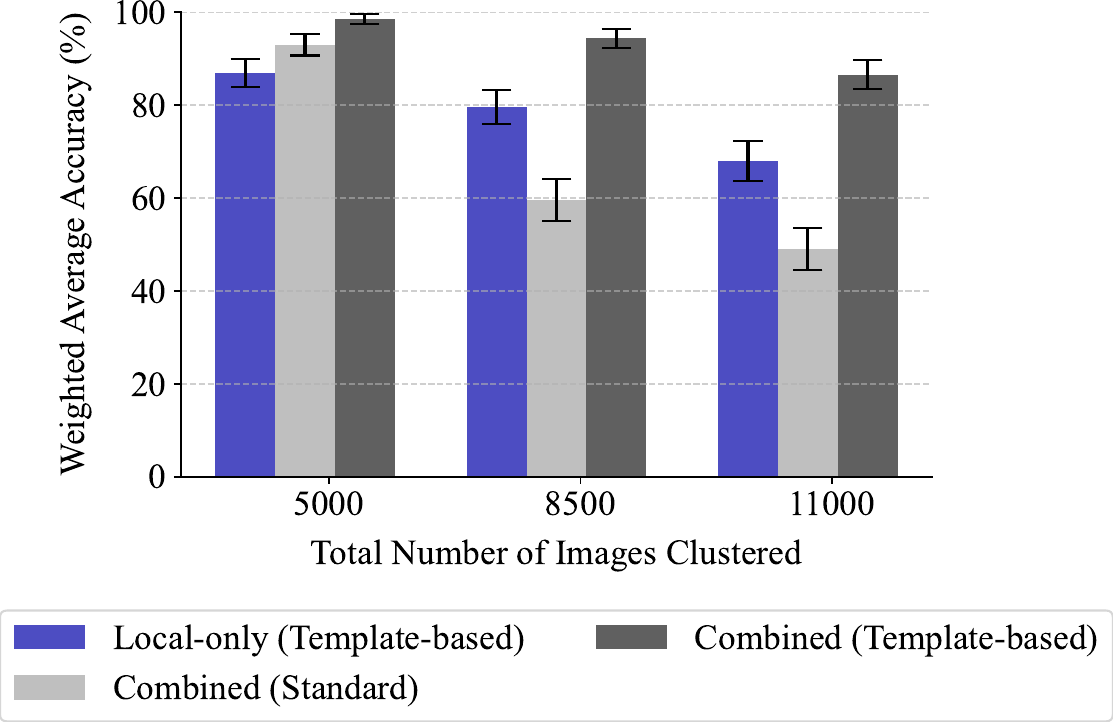}
    \caption{Mean weighted accuracy in the Imposter-Host task for various clustering methods across different numbers of total images clustered.
    Error bars indicate standard errors.}
    \label{fig:imposter-host}
\end{figure}

\noindent \textbf{Consistency with KYM clusters.}
The results in Table \ref{tab:clustering} reveal several trends. First, \textit{the template-based clustering method obtains \hl{greater} consistency across all feature sets}. Our method produces smaller and more granular clusters by centering the clustering around the identified meme templates. Instead, the clusters produced by the standard clustering baselines are often based on superficial similarities that do not align semantically with the underlying meme templates. The gap between standard \cite{theisen.2022} and template-based clustering \hl{increases} when more memes are clustered. Our template-based clustering retains high consistency when all 11,000 memes are clustered, whereas the consistency score of standard clustering drops significantly.

Second, as expected, \textit{incorporating more features leads to \hl{a} more accurate clustering}. Using all global features outperforms ViT features for both clustering methods. The combined feature set generally yields better results than relying solely on local or global features. One notable exception arises when comparing the performance of local features alone \hl{with} that of the total combined feature set under standard clustering. In this case, adding global features \hl{does not provide a} marginal benefit. However, with template-based clustering, the combined feature set significantly outperforms local features alone. This suggests that \textit{as the feature space becomes more complex, effective clustering becomes increasingly difficult without the guidance of meme templates}. When properly incorporated, global features provide a marginal but meaningful benefit. 

While the table shows these trends for Louvain clustering, we observe similar trends when using DBSCAN as a clustering algorithm, as seen in the Appendix. \hl{Further supporting these observations, an evaluation using cluster entropy to measure cluster purity (see Appendix) also indicates the superior homogeneity achieved by our template-based method. Finally, we note that the performance of CLIP is generally lower than that of ViT, suggesting that CLIP does not offer a clear advantage for meme clustering over ViT, despite its multimodal training.}

\noindent \textbf{Cluster coherence} is assessed through an Imposter-Host experiment while incrementally increasing the number of clustered images. The results for the combined and the local features are shown in Figure \ref{fig:imposter-host}. The task accuracy starts with a 93.1\% to 98.5\% score using the combined feature set at five thousand images clustered, suggesting that the identified templates are likely of reliable quality. The local features achieve a slightly lower initial accuracy of 86.5\%. 
As anticipated, the quality of the clusters \hl{decreases} as the number of clustered images increases, given the introduction of images with lower similarity scores to any template than those initially included. \textit{When comparing the template-based clustering results obtained using the combined feature set \hl{with} those derived solely from local features, we observe that the combined feature set consistently \hl{produces} higher cluster coherence at all three increments}. The marginal benefits of \textit{combining more features become more \hl{evident} at higher increments}. At 11,000 images clustered, the combined features obtain an accuracy of 86.5\% compared to 67.98\% for the local features only. \hl{In particular}, the accuracy of the combined feature set here matches that of the local-only features when clustering 5,000 images, highlighting its strength.

The figure also shows \hl{that} standard clustering lags significantly behind the template-based approach (accuracy of 49.1\% for 11,000 memes), unable to \hl{fully} exploit its extensive feature set. These findings 
generalize the results in Table \ref{tab:clustering} to \hl{an evaluation on} an expanded dataset that includes Reddit memes \hl{and contains explicit human judgments}.

\subsection{Analysis of Similarity Dimensions}
\label{matches}

\noindent \textbf{Impact of similarity-based feature sets.}
Figure \ref{fig:additional_clustered} presents the results for our four similarity feature sets and the combined set of \hl{the} six features. The combined feature set consistently obtains the highest or second-highest accuracy, highlighting the benefits of integrating various similarity dimensions to form more reliable meme clusters. At higher increments of memes matched, accuracy rates remain considerably higher than those of any individual similarity dimension. This allows us to cluster many memes that do not follow clear template structures with relatively high precision, addressing the gap in~\cite{joshi.et.al.2023}. 

Among the individual feature sets, clustering based on \textit{identity} exhibits a high accuracy, especially for the initial 2,000 matched memes, but experiences a sharp decline afterward. This highlights the discrete nature of identity similarity: images depict the same person or do not, making matches in this dimension entirely accurate or false. The only exception is group photos with partial overlap \hl{with respect to} the individuals present.

\begin{figure}[!t]
    \centering
    \includegraphics[width=0.88\linewidth]{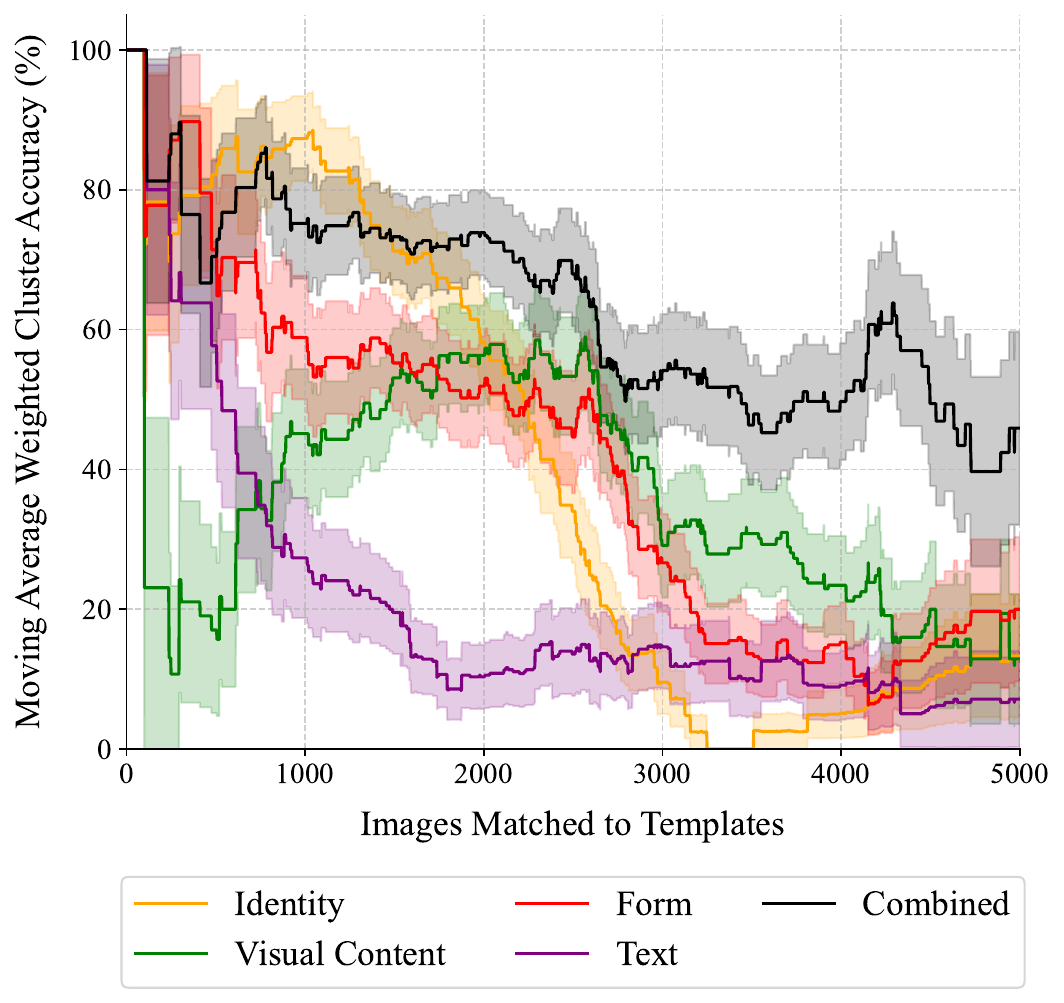}
    \caption{Moving average accuracy of image matching using various feature sets. At each increment of additional images matched, the average is estimated over all clusters with that many images matched or fewer, using a rolling window.
    Highlighted areas represent the standard error of the mean.}
    \label{fig:additional_clustered}
\end{figure}

\begin{figure}[!t]
    \centering
    \includegraphics[width=0.92\linewidth]{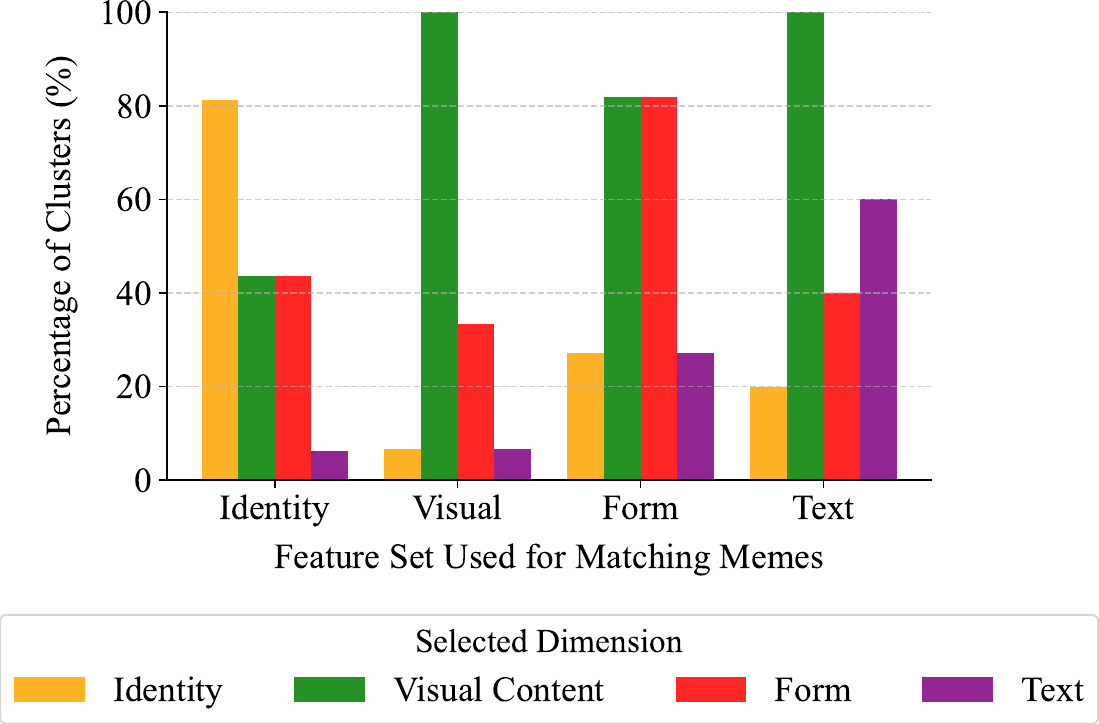}
    \caption{The percentage of clusters, deemed accurate by humans, for which various similarity dimensions were selected in response to: 'Select all the ways in which the memes [in this cluster] are related.'}
    \label{fig:freq_feature_sets}
\end{figure}

\begin{figure*}[!t]
    \centering
    \includegraphics[width=0.97\linewidth]{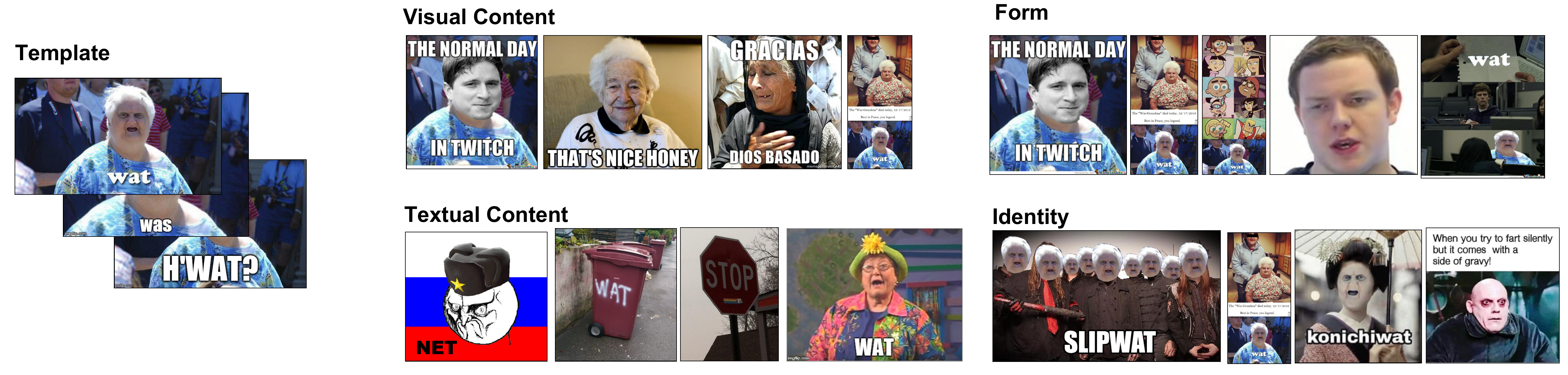}
    \caption{Identified template for `WAT Grandma' meme and matches through various feature dimensions, ordered by similarity.}
    \label{fig:example_case}
\end{figure*}

The performance of the \textit{form} shows a different pattern, with accuracy dropping off quickly before stabilizing somewhat and declining again. This aligns with the inherent limitations of similarity assessment based solely on the form: While algorithms may accurately identify shared local interest points or color histograms within images, these commonalities do not necessarily denote semantic relevance. For instance, a significant proportion of memes originate as screenshots, inheriting common user interface elements such as repost buttons that do not signify a meaningful relationship between these memes themselves, rendering them inadequate as criteria for accurate matching. 

\hl{The} matching based on \textit{visual content} exhibits a \hl{trend similar} to that observed for the form-based approach, except with an even sharper drop at the \hl{beginning}. Part of this is due to the influence of outliers, which carry more weight early on since there are fewer data points in the moving average. Upon manual inspection, we discovered that the early incorrect matches were all screenshots with minimal visual content, such as text-only tweets. Although precision is low, the performance of visual content matching remains relatively high at higher increments, suggesting that the features can still intermittently yield relevant matches.

\textit{Text-based} matching performs the poorest. Its accuracy declines steadily after clustering just a few additional images, eventually plateauing \hl{at a} very low \hl{accuracy}. This result is intuitive, as memes often share visual elements while using textual elements for \hl{individuality}. Only a handful of meme clusters rely heavily on shared textual elements, such as catchphrases. Even when these elements are present, accurate matching is challenging because they often appear as fill-in-the-blank templates combined with other text.

\noindent \textbf{Alignment with human similarity labels.}
To validate whether the employed feature sets correctly target the intended similarity dimensions, human judges were instructed to identify the dimensions through which memes are related for clusters they rated accurate. Figure \ref{fig:freq_feature_sets} shows the frequency with which various dimensions were selected for matches based on each distinct feature set. \textit{The feature sets tend to successfully capture the intended aspect of meme similarity, as the most commonly selected dimension aligns with the target for each feature set}. The exception is \hl{the textual feature set}, \hl{which reveals} that the annotators may occasionally \hl{have} overlooked meme text, focusing more on visual similarities. 

\hl{Notably, Figure \ref{fig:freq_feature_sets} shows that the matching based on form often also results in high visual content similarity. However, this is expected: images that are pixel-wise similar typically depict similar subjects. Moreover, we have noticed that aesthetic similarity alone, e.g., a shared color palette, rarely leads people to judge memes as related unless another similarity dimension is also present. This suggests that form, on its own, is a relatively weak signal. The figure also indicates that when images are matched according to content, in only about 30\% of cases, annotators consider their form to be similar as well. This supports our hypothesis that \textit{form and content are meaningfully distinct in how we measure them}: visual content similarity does not simply arise from low-level visual resemblance. This also confirms our choice of ViT- over CLIP-based models to model the similarity dimensions of Internet memes.}

\subsection{Case Studies}

\noindent \textbf{Case 1: ``WAT Grandma''}\footnote{\url{https://knowyourmeme.com/memes/wat}} is presented in Figure \ref{fig:example_case} together with memes similar to it according to various dimensions to highlight the strengths and weaknesses of our approach. This meme, featuring a confused elderly woman with the caption ``wat'', humorously depicts bewilderment in response to something nonsensical or hard to understand.

\begin{figure*}[!t]
    \centering
    \includegraphics[width=0.97\linewidth]{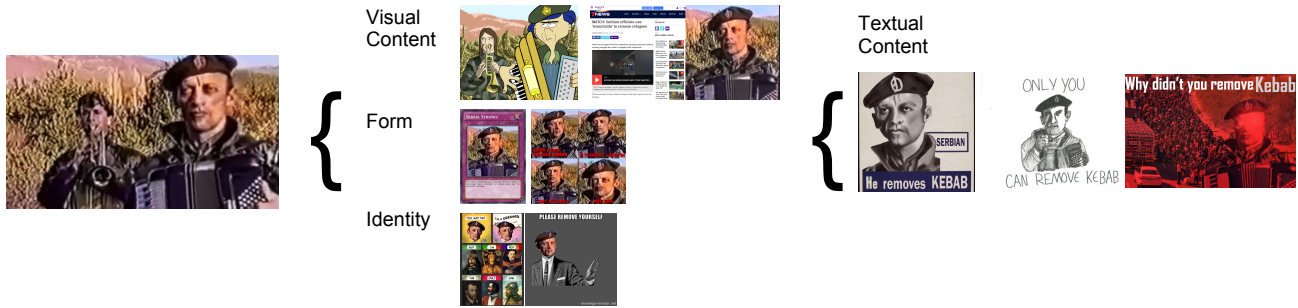}
    \caption{The `Remove Kebab' meme and its matches through various similarity dimensions.}
    \label{fig:appendix_kebab}
\end{figure*}

When examining \textit{visual} content, we note that multiple memes containing images of elderly women are matched to the ``WAT Grandma" template, even if they do not adhere to the original meme's specific visual structure or textual elements. Although these memes are related, they do not capture the core semantics of the template, focusing on the woman's age rather than her confused look. 
We observe similar patterns for matches based solely on \textit{form} features.
Although demonstrating the lowest overall accuracy in our findings, \textit{textual} features effectively capture the semantic link between the \hl{template} ``WAT Grandma"  and other memes containing the ``wat" caption. This finding highlights the importance of incorporating textual content into meme analysis, as it can reveal connections that might be missed by visual features alone. However, even in the case of this straightforward catchphrase, text-based matching reveals relatively low accuracy, as evidenced by two erroneous matches that scored higher in similarity than at least one correct match. Text-based similarity is challenging for memes, as errors may arise when extracting text from images and embedding them to emphasize specific elements\hl{, such as} phrasal templates, rather than the overall semantic content.
Conversely, the \textit{identity} features successfully link the ``WAT Grandma" template to other memes featuring the same person, even if these memes deviate from the template's typical form. This underscores the value of identity as a similarity dimension, mainly for memes centered around individuals.

The \textit{combined} feature leverages the complementarity of these four dimensions by aggregating their scores for more effective clustering.
For \hl{example}, the fourth match based on the \textit{form} dimension may be prioritized over the erroneous third match because it also exhibits similarity through \textit{text} and \textit{identity}. 
This example also highlights two limitations of our method. First, the distribution of similarity scores across features is essential, which makes it especially challenging to assign high scores to memes that only rank high in one similarity dimension.
Second, our approach remains sensitive to the quality and focus of the selected feature extractors that model a similarity dimension.

\noindent \hl{\textbf{Case 2: ``Remove kebab''}\footnote{\url{https://knowyourmeme.com/memes/serbia-strong-remove-kebab}} demonstrates the applicability of our methodology in the context of hateful and toxic memes (Figure \ref{fig:appendix_kebab}). This template originates from a screenshot of a Serbian nationalist, anti-Croat, and anti-Muslim propaganda music video from the Yugoslav Wars, which has gained notoriety through viral spread among far-right nationalist groups. We demonstrate how related meme instances are matched through distinct similarity dimensions. We present examples selected among matches with the top ten highest similarity scores identified using the feature sets specifically designed for each dimension.}

\hl{Employing the \textit{visual content} dimension, our method identifies memes such as a comic-style illustration derived from the original screenshot; notably, this match lacks direct pixel correspondence, highlighting the necessity of this dimension for capturing such variations. Matches based on \textit{form} similarity retrieve memes that share substantial pixel-level visual overlap with the template, even when overlaid by other visual elements. The \textit{identity} dimension, leveraging facial features, isolates instances where only the face of the Bosnian Serb Army soldier featured in the source video is retained. At the same time, the context of the surrounding image is altered or removed.}

\hl{These examples underscore the utility of analyzing distinct dimensions. As certain relevant matches are discoverable \textit{only} through a specific dimension, omitting these dimensions would consequently result in lower recall. Furthermore, while the original screenshot lacks overlaid text, memes matched via visual similarity dimensions (identity, form, and visual content) frequently exhibit recurring keywords and phrases associated with the meme's hateful origins, such as `Serb', `Kebab', and `Remove'. If these visually similar instances form a coherent cluster identified as a template in our process, subsequent matching could leverage the \textit{textual content} dimension to find further related memes containing these phrases, as conceptually illustrated on the right of Figure \ref{fig:appendix_kebab}. Although not illustrated, using the \textit{ combined} feature set offers the potential for even more in-depth image matching. For instance, the co-occurrence of the phrase `kebab' and imagery related to soldiers constitutes a stronger signal that a meme instance is related to this template than either feature would individually, potentially enabling the matching of more subtly related memes.}

\hl{This case study indicates the need for a comprehensive, multi-dimensional perspective when analyzing potentially toxic or harmful memes. The subtle nature of memetic communication, including obfuscated references to problematic origins, can evade detection by methods that rely solely on standard computer vision techniques unless similarity is assessed and integrated across multiple dimensions, as our framework facilitates. Meanwhile, to apply our methodology meaningfully to detect toxic memes, it is necessary to design a toxicity scoring mechanism for the identified templates and use those scores to infer the toxicity of individual memes. This extension of our method, in line with \cite{bates-et-al-2023}, is a promising future direction.}

\section{Conclusion}

This paper contributes a modular, multi-dimensional approach to meme clustering that goes beyond simple visual similarity analysis. Our method leverages automatically identified templates and diverse global and local features to capture meme similarity across form, content, and identity. \hl{The grounding of} the clustering in templates aligns with the inherent structure of memes, which often involve variations on a shared semantic basis.
Experiments with curated meme databases and human judgments show that our multi-dimensional approach yields consistent, coherent, and semantically meaningful clusters. 
Further quantitative and qualitative analysis shows the complementary nature of individual feature sets and their alignment with human similarity labels.

\hl{Our two-step strategy \textemdash first clustering a subset of highly similar memes to discover coherent templates, then matching additional memes incrementally \textemdash proves crucial in avoiding overly broad, noisy clusters.} By addressing the complex nature of memes and their relationships, our method enables more nuanced and accurate analyses of these popular artifacts for digital communication, which can be further assisted by
similarity-aware semantic search engines for memes, as illustrated in the Appendix.
Our method thus provides a balanced and systematic perspective on studying the toxicity and semantics of Internet memes by social scientists and content moderators.

Our method remains sensitive to the quality and interplay of the feature extraction methods. \hl{In addition}, it does not incorporate background knowledge about culture, personal values, or intent. 
Future research should increase the robustness of similarity features, devise methods to explicitly model the interplay between these dimensions, and incorporate background knowledge from sources \hl{such as} IMKG and ConceptNet. 
Future work should apply our clustering approach to the downstream tasks \hl{such as tracking} meme evolution, sentiment analysis, and content moderation. 
While our work aims to support social scientists and content moderators toward a safer social media environment, we acknowledge that advances in meme interpretation can be misused to harm individuals or social groups. Our public release of the code enables responsible use of our method, yet further strategies are necessary to mitigate its misuse.

\section{Acknowledgements}

This paper is based on Tygo Bloem's MSc AI project at Vrije Universiteit Amsterdam, supervised by Filip Ilievski. The authors thank the reviewers for their insightful suggestions, the data annotators for their diligent support, and the second thesis assessor, Victor de Boer, for his feedback.

\bibliography{aaai25}

\begin{thebibliography}{39}
\providecommand{\natexlab}[1]{#1}

\bibitem[{Bates et~al.(2023)Bates, Christensen, Nakov, and Gurevych}]{bates-et-al-2023}
Bates, L.; Christensen, P.~E.; Nakov, P.; and Gurevych, I. 2023.
\newblock A Template Is All You Meme.
\newblock arXiv:2311.06649.

\bibitem[{Bay et~al.(2008)Bay, Ess, Tuytelaars, and {Van Gool}}]{bay.et.al.2006}
Bay, H.; Ess, A.; Tuytelaars, T.; and {Van Gool}, L. 2008.
\newblock Speeded-Up Robust Features (SURF).
\newblock \emph{Computer Vision and Image Understanding}, 110(3): 346--359.
\newblock Similarity Matching in Computer Vision and Multimedia.

\bibitem[{Beskow, Kumar, and Carley(2020)}]{beskow.et.al.2020}
Beskow, D.~M.; Kumar, S.; and Carley, K.~M. 2020.
\newblock The evolution of political memes: Detecting and characterizing internet memes with multi-modal deep learning.
\newblock \emph{Information Processing and Management}, 57(2): 102170.

\bibitem[{Blondel et~al.(2008)Blondel, Guillaume, Lambiotte, and Lefebvre}]{Blondel_2008}
Blondel, V.~D.; Guillaume, J.-L.; Lambiotte, R.; and Lefebvre, E. 2008.
\newblock Fast unfolding of communities in large networks.
\newblock \emph{Journal of Statistical Mechanics: Theory and Experiment}, 2008(10): P10008.

\bibitem[{Dancygier and Vandelanotte(2017)}]{Dancygier.Vandelanotte.2017}
Dancygier, B.; and Vandelanotte, L. 2017.
\newblock Internet memes as multimodal constructions.
\newblock \emph{Cognitive Linguistics}, 28(3): 565--598.

\bibitem[{Devlin et~al.(2019)Devlin, Chang, Lee, and Toutanova}]{devlin2019}
Devlin, J.; Chang, M.-W.; Lee, K.; and Toutanova, K. 2019.
\newblock BERT: Pre-training of Deep Bidirectional Transformers for Language Understanding.
\newblock arXiv:1810.04805.

\bibitem[{Dosovitskiy et~al.(2021)Dosovitskiy, Beyer, Kolesnikov, Weissenborn, Zhai, Unterthiner, Dehghani, Minderer, Heigold, Gelly, Uszkoreit, and Houlsby}]{Dosovitskiy.et.al.2020}
Dosovitskiy, A.; Beyer, L.; Kolesnikov, A.; Weissenborn, D.; Zhai, X.; Unterthiner, T.; Dehghani, M.; Minderer, M.; Heigold, G.; Gelly, S.; Uszkoreit, J.; and Houlsby, N. 2021.
\newblock An Image is Worth 16x16 Words: Transformers for Image Recognition at Scale.
\newblock In \emph{9th International Conference on Learning Representations, {ICLR} 2021, Virtual Event, Austria, May 3-7, 2021}. OpenReview.net.

\bibitem[{Dubey et~al.(2018)Dubey, Moro, Cebrian, and Rahwan}]{dubey.et.al.2018}
Dubey, A.; Moro, E.; Cebrian, M.; and Rahwan, I. 2018.
\newblock MemeSequencer: Sparse Matching for Embedding Image Macros.
\newblock In \emph{Proceedings of the 2018 World Wide Web Conference}, WWW '18, 1225–1235. Republic and Canton of Geneva, CHE: International World Wide Web Conferences Steering Committee.
\newblock ISBN 9781450356398.

\bibitem[{{FORCE11}(2020)}]{fair}
{FORCE11}. 2020.
\newblock The FAIR Data principles.
\newblock \url{https://force11.org/info/the-fair-data-principles/}.
\newblock Accessed: 2025-04-24.

\bibitem[{Gebru et~al.(2021)Gebru, Morgenstern, Vecchione, Vaughan, Wallach, Iii, and Crawford}]{gebru2021datasheets}
Gebru, T.; Morgenstern, J.; Vecchione, B.; Vaughan, J.~W.; Wallach, H.; Iii, H.~D.; and Crawford, K. 2021.
\newblock Datasheets for datasets.
\newblock \emph{Communications of the ACM}, 64(12): 86--92.

\bibitem[{Grasso et~al.(2024)Grasso, {La Gatta}, Moscato, and Sperlì}]{Grasso-et-al-2024}
Grasso, B.; {La Gatta}, V.; Moscato, V.; and Sperlì, G. 2024.
\newblock KERMIT: Knowledge-EmpoweRed Model In harmful meme deTection.
\newblock \emph{Information Fusion}, 106: 102269.

\bibitem[{Groppe and Jain(2024)}]{groppe2024way}
Groppe, S.; and Jain, S. 2024.
\newblock The Way Forward with AI-Complete Problems.
\newblock \emph{New Generation Computing}, 42(1): 1--5.

\bibitem[{Hee, Chong, and Lee(2023)}]{hee.et.al.2023}
Hee, M.~S.; Chong, W.-H.; and Lee, R. K.-W. 2023.
\newblock Decoding the underlying meaning of multimodal hateful memes.
\newblock In \emph{Proceedings of the Thirty-Second International Joint Conference on Artificial Intelligence}, IJCAI '23.
\newblock ISBN 978-1-956792-03-4.

\bibitem[{Howard et~al.(2019)Howard, Sandler, Chu, Chen, Chen, Tan, Wang, Zhu, Pang, Vasudevan et~al.}]{howard2019searching}
Howard, A.; Sandler, M.; Chu, G.; Chen, L.-C.; Chen, B.; Tan, M.; Wang, W.; Zhu, Y.; Pang, R.; Vasudevan, V.; et~al. 2019.
\newblock Searching for mobilenetv3.
\newblock In \emph{Proceedings of the IEEE/CVF international conference on computer vision}, 1314--1324.

\bibitem[{Johnson, Douze, and J{\'e}gou(2019)}]{faiss}
Johnson, J.; Douze, M.; and J{\'e}gou, H. 2019.
\newblock Billion-scale similarity search with {GPUs}.
\newblock \emph{IEEE Transactions on Big Data}, 7(3): 535--547.

\bibitem[{Joshi, Ilievski, and Luceri(2023)}]{joshi.et.al.2023}
Joshi, S.; Ilievski, F.; and Luceri, L. 2023.
\newblock Contextualizing Internet Memes Across Social Media Platforms.
\newblock \emph{Companion Proceedings of the ACM on Web Conference 2024}.

\bibitem[{Kiela et~al.(2020)Kiela, Firooz, Mohan, Goswami, Singh, Ringshia, and Testuggine}]{kiela2021hateful}
Kiela, D.; Firooz, H.; Mohan, A.; Goswami, V.; Singh, A.; Ringshia, P.; and Testuggine, D. 2020.
\newblock The hateful memes challenge: detecting hate speech in multimodal memes.
\newblock In \emph{Proceedings of the 34th International Conference on Neural Information Processing Systems}, NIPS '20. Red Hook, NY, USA: Curran Associates Inc.
\newblock ISBN 9781713829546.

\bibitem[{King(2009)}]{dlib}
King, D.~E. 2009.
\newblock Dlib C++ Library.
\newblock \url{http://dlib.net}.
\newblock Accessed: 2024-06-24.

\bibitem[{Klinger and Starkweather(2013)}]{klinger2013phash}
Klinger, E.; and Starkweather, D. 2013.
\newblock {pHash: The open source perceptual hash library}.
\newblock \url{https://www.phash.org}.
\newblock Accessed: 2024-06-23.

\bibitem[{Liu et~al.(2022)Liu, Geigle, Krebs, and Gurevych}]{liu-etal-2022-figmemes}
Liu, C.; Geigle, G.; Krebs, R.; and Gurevych, I. 2022.
\newblock {F}ig{M}emes: A Dataset for Figurative Language Identification in Politically-Opinionated Memes.
\newblock In Goldberg, Y.; Kozareva, Z.; and Zhang, Y., eds., \emph{Proceedings of the 2022 Conference on Empirical Methods in Natural Language Processing}, 7069--7086. Abu Dhabi, United Arab Emirates: Association for Computational Linguistics.

\bibitem[{Lowe(1999)}]{lowe.1999}
Lowe, D. 1999.
\newblock Object recognition from local scale-invariant features.
\newblock In \emph{Proceedings of the Seventh IEEE International Conference on Computer Vision}, volume~2, 1150--1157 vol.2.

\bibitem[{Morinan(2021)}]{morinan-2021}
Morinan, G. 2021.
\newblock {Meme Vision: the science of classifying memes}.
\newblock \url{https://towardsdatascience.com/meme-vision-framework-e90a9a7a4187}.
\newblock Accessed: 2025-04-15.

\bibitem[{Onielfa, Casacuberta, and Escalera(2022)}]{Onielfa.et.al.2022}
Onielfa, C.; Casacuberta, C.; and Escalera, S. 2022.
\newblock Influence in Social Networks Through Visual Analysis of Image Memes.
\newblock In \emph{Artificial Intelligence Research and Development}, 71--80. IOS Press.

\bibitem[{Pandiani, Sang, and Ceolin(2024)}]{pandiani2024toxic}
Pandiani, D. S.~M.; Sang, E. T.~K.; and Ceolin, D. 2024.
\newblock Toxic Memes: A Survey of Computational Perspectives on the Detection and Explanation of Meme Toxicities.
\newblock \emph{arXiv preprint arXiv:2406.07353}.

\bibitem[{Peirson and Tolunay(2018)}]{perison.tolunay.2018}
Peirson, A.~L.; and Tolunay, E.~M. 2018.
\newblock Dank learning: Generating memes using deep neural networks.
\newblock \emph{arXiv preprint arXiv:1806.04510}.

\bibitem[{Pramanick et~al.(2021)Pramanick, Dimitrov, Mukherjee, Sharma, Akhtar, Nakov, and Chakraborty}]{Pramanick.et.al.2021}
Pramanick, S.; Dimitrov, D.; Mukherjee, R.; Sharma, S.; Akhtar, M.~S.; Nakov, P.; and Chakraborty, T. 2021.
\newblock Detecting Harmful Memes and Their Targets.
\newblock In Zong, C.; Xia, F.; Li, W.; and Navigli, R., eds., \emph{Findings of the Association for Computational Linguistics: ACL-IJCNLP 2021}, 2783--2796. Online: Association for Computational Linguistics.

\bibitem[{Qu et~al.(2023)Qu, He, Pierson, Backes, Zhang, and Zannettou}]{qu.et.al.2023}
Qu, Y.; He, X.; Pierson, S.; Backes, M.; Zhang, Y.; and Zannettou, S. 2023.
\newblock On the Evolution of (Hateful) Memes by Means of Multimodal Contrastive Learning.
\newblock In \emph{2023 IEEE Symposium on Security and Privacy (SP)}, 293--310.

\bibitem[{Sharma et~al.(2022)Sharma, Suresh, Kulkarni, Mathur, Nakov, Akhtar, and Chakraborty}]{Sharma_2022}
Sharma, S.; Suresh, T.; Kulkarni, A.; Mathur, H.; Nakov, P.; Akhtar, M.~S.; and Chakraborty, T. 2022.
\newblock Findings of the {CONSTRAINT} 2022 {Shared} {Task} on {Detecting} the {Hero}, the {Villain}, and the {Victim} in {Memes}.
\newblock In Chakraborty, T.; Akhtar, M.~S.; Shu, K.; Bernard, H.~R.; Liakata, M.; Nakov, P.; and Srivastava, A., eds., \emph{Proceedings of the {Workshop} on {Combating} {Online} {Hostile} {Posts} in {Regional} {Languages} during {Emergency} {Situations}}, 1--11. Dublin, Ireland: Association for Computational Linguistics.

\bibitem[{Shifman(2019)}]{shifman2019internet}
Shifman, L. 2019.
\newblock Internet memes and the twofold articulation of values.
\newblock \emph{Society and the internet: How networks of information and communication are changing our lives}, 43--57.

\bibitem[{Simonyan and Zisserman(2015)}]{simonyan2014very}
Simonyan, K.; and Zisserman, A. 2015.
\newblock Very Deep Convolutional Networks for Large-Scale Image Recognition.
\newblock In Bengio, Y.; and LeCun, Y., eds., \emph{3rd International Conference on Learning Representations, {ICLR} 2015, San Diego, CA, USA, May 7-9, 2015, Conference Track Proceedings}.

\bibitem[{Thakur et~al.(2023)Thakur, Ilievski, Sandlin, Sourati, Luceri, Tommasini, and Mermoud}]{thakur2023multimodal}
Thakur, A.~K.; Ilievski, F.; Sandlin, H.~{\^A}.; Sourati, Z.; Luceri, L.; Tommasini, R.; and Mermoud, A. 2023.
\newblock Explainable Classification of Internet Memes.
\newblock In \emph{17th International Workshop on Neural-Symbolic Learning and Reasoning, NeSy 2023}.

\bibitem[{Theisen et~al.(2021)Theisen, Brogan, Thomas, Moreira, Phoa, Weninger, and Scheirer}]{theisen.et.al.2020}
Theisen, W.; Brogan, J.; Thomas, P.~B.; Moreira, D.; Phoa, P.; Weninger, T.; and Scheirer, W. 2021.
\newblock Automatic discovery of political meme genres with diverse appearances.
\newblock In \emph{Proceedings of the International AAAI Conference on Web and Social Media}, volume~15, 714--726.

\bibitem[{Theisen et~al.(2023)Theisen, Cedre, Carmichael, Moreira, Weninger, and Scheirer}]{theisen.2022}
Theisen, W.; Cedre, D.~G.; Carmichael, Z.; Moreira, D.; Weninger, T.; and Scheirer, W. 2023.
\newblock Motif mining: Finding and summarizing remixed image content.
\newblock In \emph{Proceedings of the IEEE/CVF Winter Conference on Applications of Computer Vision}, 1319--1328.

\bibitem[{Tommasini, Ilievski, and Wijesiriwardene(2023)}]{Tommasinietal2023}
Tommasini, R.; Ilievski, F.; and Wijesiriwardene, T. 2023.
\newblock IMKG: The Internet Meme Knowledge Graph.
\newblock In Pesquita, C.; Jimenez-Ruiz, E.; McCusker, J.; Faria, D.; Dragoni, M.; Dimou, A.; Troncy, R.; and Hertling, S., eds., \emph{The Semantic Web}, 354--371. Cham: Springer Nature Switzerland.
\newblock ISBN 978-3-031-33455-9.

\bibitem[{Wang and Wen(2015)}]{wang-wen-2015-cheezburger}
Wang, W.~Y.; and Wen, M. 2015.
\newblock {I} Can Has Cheezburger? A Nonparanormal Approach to Combining Textual and Visual Information for Predicting and Generating Popular Meme Descriptions.
\newblock In Mihalcea, R.; Chai, J.; and Sarkar, A., eds., \emph{Proceedings of the 2015 Conference of the North {A}merican Chapter of the Association for Computational Linguistics: Human Language Technologies}, 355--365. Denver, Colorado: Association for Computational Linguistics.

\bibitem[{Weng, Menczer, and Ahn(2014)}]{Weng.et.al.2014}
Weng, L.; Menczer, F.; and Ahn, Y.-Y. 2014.
\newblock Predicting Successful Memes Using Network and Community Structure.
\newblock \emph{Proceedings of the International AAAI Conference on Web and Social Media}, 8(1): 535--544.

\bibitem[{Zannettou et~al.(2018)Zannettou, Caulfield, Blackburn, De~Cristofaro, Sirivianos, Stringhini, and Suarez-Tangil}]{Zannettou.et.al.2018}
Zannettou, S.; Caulfield, T.; Blackburn, J.; De~Cristofaro, E.; Sirivianos, M.; Stringhini, G.; and Suarez-Tangil, G. 2018.
\newblock On the Origins of Memes by Means of Fringe Web Communities.
\newblock In \emph{Proceedings of the Internet Measurement Conference 2018}, IMC '18, 188–202. New York, NY, USA: Association for Computing Machinery.
\newblock ISBN 9781450356190.

\bibitem[{Zhou, Jurgens, and Bamman(2024)}]{zhou-etal-2024-social}
Zhou, N.; Jurgens, D.; and Bamman, D. 2024.
\newblock Social Meme-ing: Measuring Linguistic Variation in Memes.
\newblock In Duh, K.; Gomez, H.; and Bethard, S., eds., \emph{Proceedings of the 2024 Conference of the North American Chapter of the Association for Computational Linguistics: Human Language Technologies (Volume 1: Long Papers)}, 3005--3024. Mexico City, Mexico: Association for Computational Linguistics.

\bibitem[{Zhou et~al.(2017)Zhou, Yao, Wen, Wang, Zhou, He, and Liang}]{DBLP:journals/corr/ZhouYWWZHL17}
Zhou, X.; Yao, C.; Wen, H.; Wang, Y.; Zhou, S.; He, W.; and Liang, J. 2017.
\newblock {EAST:} an efficient and accurate scene text detector.
\newblock In \emph{Proceedings of the IEEE conference on Computer Vision and Pattern Recognition}, 5551--5560.

\end{thebibliography}

\clearpage
\newpage
\section{Paper Checklist}

\begin{enumerate}

\item For most authors...
\begin{enumerate}
    \item  Would answering this research question advance science without violating social contracts, such as violating privacy norms, perpetuating unfair profiling, exacerbating the socio-economic divide, or implying disrespect to societies or cultures?
    Yes, the multi-dimensional similarity perspective of this work enables for nuanced matching of memes.
  \item Do your main claims in the abstract and introduction accurately reflect the paper's contributions and scope?
    Yes, the research question and the contributions listed at the end of the Introduction section are aligned with the title, abstract, methodology, and results.
   \item Do you clarify how the proposed methodological approach is appropriate for the claims made? 
    Yes, the Methodology section supports contributions 1 and 2; the Experimental Setup defines how the method is evaluated; and the Results section supports contribution 3.
   \item Do you clarify what are possible artifacts in the data used, given population-specific distributions?
    Yes, Data Details and Human Validation Details and Interface (Appendix) report information about data considerations, and the size and the demographics of the human judges.
  \item Did you describe the limitations of your work?
    Yes, the limitations of our method are presented at the end of the Results section (last paragraph). Further limitations, coupled with possible mitigation strategies, are presented in the second paragraph of the Conclusion section.
  \item Did you discuss any potential negative societal impacts of your work?
    Yes, the societal impact of meme similarity is captured in the Introduction section, and we acknowledge the possible negative use of our method at the end of the Conclusion.
      \item Did you discuss any potential misuse of your work?
    Yes,  we acknowledge the possible misuse of our method at the end of the Conclusion.
    \item Did you describe steps taken to prevent or mitigate potential negative outcomes of the research, such as data and model documentation, data anonymization, responsible release, access control, and the reproducibility of findings?
    Yes, see the end of the Conclusion, as well as the Implementation Details and the Data Details in the Appendix.
  \item Have you read the ethics review guidelines and ensured that your paper conforms to them?
    Yes.
\end{enumerate}

\item Additionally, if your study involves hypotheses testing...
\begin{enumerate}
  \item Did you clearly state the assumptions underlying all theoretical results?
  Yes, the Methodology section captures the nuances in the design of our framework.
  \item Have you provided justifications for all theoretical results?
    Yes, the motivation for splitting features into local/global and into similarity dimensions is described in the Methodology and the Related Work.
  \item Did you discuss competing hypotheses or theories that might challenge or complement your theoretical results?
    Yes, see Methodology as well as the design of the baselines in Experimental Setup.
  \item Have you considered alternative mechanisms or explanations that might account for the same outcomes observed in your study?
    Yes, we experiment with multiple tasks and alternative framework components (e.g., two clustering algorithms), see Experimental Setup. Moreover, we provide ablation studies to rule out alternative explanations to a reasonable extent, see Results.
  \item Did you address potential biases or limitations in your theoretical framework?
   Yes, we acknowledge the fact that our set of similarity dimensions may not be the optimal one, see Adjacency Matrix Construction.
  \item Have you related your theoretical results to the existing literature in social science?
    Yes, we ground our separation of local and global features, and our analysis of similarity, into prior work, see Related Work.
  \item Did you discuss the implications of your theoretical results for policy, practice, or further research in the social science domain?
    Yes, see Conclusions.
\end{enumerate}

\item Additionally, if you are including theoretical proofs...
\begin{enumerate}
  \item Did you state the full set of assumptions of all theoretical results?
    NA.
	\item Did you include complete proofs of all theoretical results?
    NA.
\end{enumerate}

\item Additionally, if you ran machine learning experiments...
\begin{enumerate}
  \item Did you include the code, data, and instructions needed to reproduce the main experimental results (either in the supplemental material or as a URL)?
    Yes, we included the entire code of our method. Upon acceptance, we will provide comprehensive documentation about running the code, including instructions on how to obtain the data.
  \item Did you specify all the training details (e.g., data splits, hyperparameters, how they were chosen)?
    Yes, please see Implementation Details in the Appendix.
     \item Did you report error bars (e.g., with respect to the random seed after running experiments multiple times)?
    No, because we noticed the experimental results had only a slight variance across runs.
	\item Did you include the total amount of compute and the type of resources used (e.g., type of GPUs, internal cluster, or cloud provider)?
    Yes, please see Implementation Details.
     \item Do you justify how the proposed evaluation is sufficient and appropriate to the claims made? 
    Yes, see Experimental Setup.
     \item Do you discuss what is ``the cost`` of misclassification and fault (in)tolerance?
    No, because our method enables a multi-dimensional approach to similarity-based classification of memes, which alleviates the need for a single best classification.
  
\end{enumerate}

\item Additionally, if you are using existing assets (e.g., code, data, models) or curating/releasing new assets, \textbf{without compromising anonymity}...
\begin{enumerate}
  \item If your work uses existing assets, did you cite the creators?
    Yes, see Data in Experimental Setup for the data sources we use and the Implementation Details in the Appendix for code details.
  \item Did you mention the license of the assets?
    Yes, see Implementation Details and Data Details in the Appendix for the code and the data, respectively.
  \item Did you include any new assets in the supplemental material or as a URL?
    Yes, we include our experimental code in the uploaded supplementary material.
  \item Did you discuss whether and how consent was obtained from people whose data you're using/curating?
    Yes, see Data Details and Implementation Details in the Appendix.
  \item Did you discuss whether the data you are using/curating contains personally identifiable information or offensive content?
    Yes, see Conclusion: the data does not contain personally identifiable information, whereas some offensive content may be part of the data.
\item If you are curating or releasing new datasets, did you discuss how you intend to make your datasets FAIR (see \citet{fair})?
NA.
\item If you are curating or releasing new datasets, did you create a Datasheet for the Dataset (see \citet{gebru2021datasheets})? 
NA.
\end{enumerate}

\item Additionally, if you used crowdsourcing or conducted research with human subjects, \textbf{without compromising anonymity}...
\begin{enumerate}
  \item Did you include the full text of instructions given to participants and screenshots?
    Yes, please see Human Validation Details and Interface.
  \item Did you describe any potential participant risks, with mentions of Institutional Review Board (IRB) approvals?
    NA.
  \item Did you include the estimated hourly wage paid to participants and the total amount spent on participant compensation?
    The human judges participated voluntarily and were not financially compensated.
   \item Did you discuss how data is stored, shared, and deidentified?
   Yes, see Data Details in the Appendix.
\end{enumerate}

\end{enumerate}
\appendix

\section{Appendix}
\label{AppendixA}

\subsection{Human Validation Details and Interface}

All 12 human judges are adults in higher education or have completed a higher education track. \hl{Almost} all come from the Netherlands and are adults between \hl{the ages of} 18 and 64.
Each human was presented with 90 meme clusters for the Imposter-Host task and 50 for the second judgment task.


The following figures illustrate the interfaces used by human judges for the manual evaluation process.
Before \hl{performing} the task, human judges receive \hl{the} instructions presented in Figure \ref{fig:instructions}.
Figure \ref{fig:imposter-host-setup} presents an example with two clusters. A red outline highlights the images identified by the user as imposters. \hl{The} images can be enlarged by clicking the button located in the top-left corner. 
Figure \ref{fig:meme-validation-setup} illustrates our validation task with two clusters. Users are prompted to respond to the question "Are all the memes above related?" by clicking either the Yes or No button. When No is selected, a pop-up prompts the user to specify how the memes are related. Users can also enlarge the images by clicking on them.

\begin{figure}[!ht]
    \begin{subfigure}[b]{\linewidth}
        \centering
        \includegraphics[width=\linewidth]{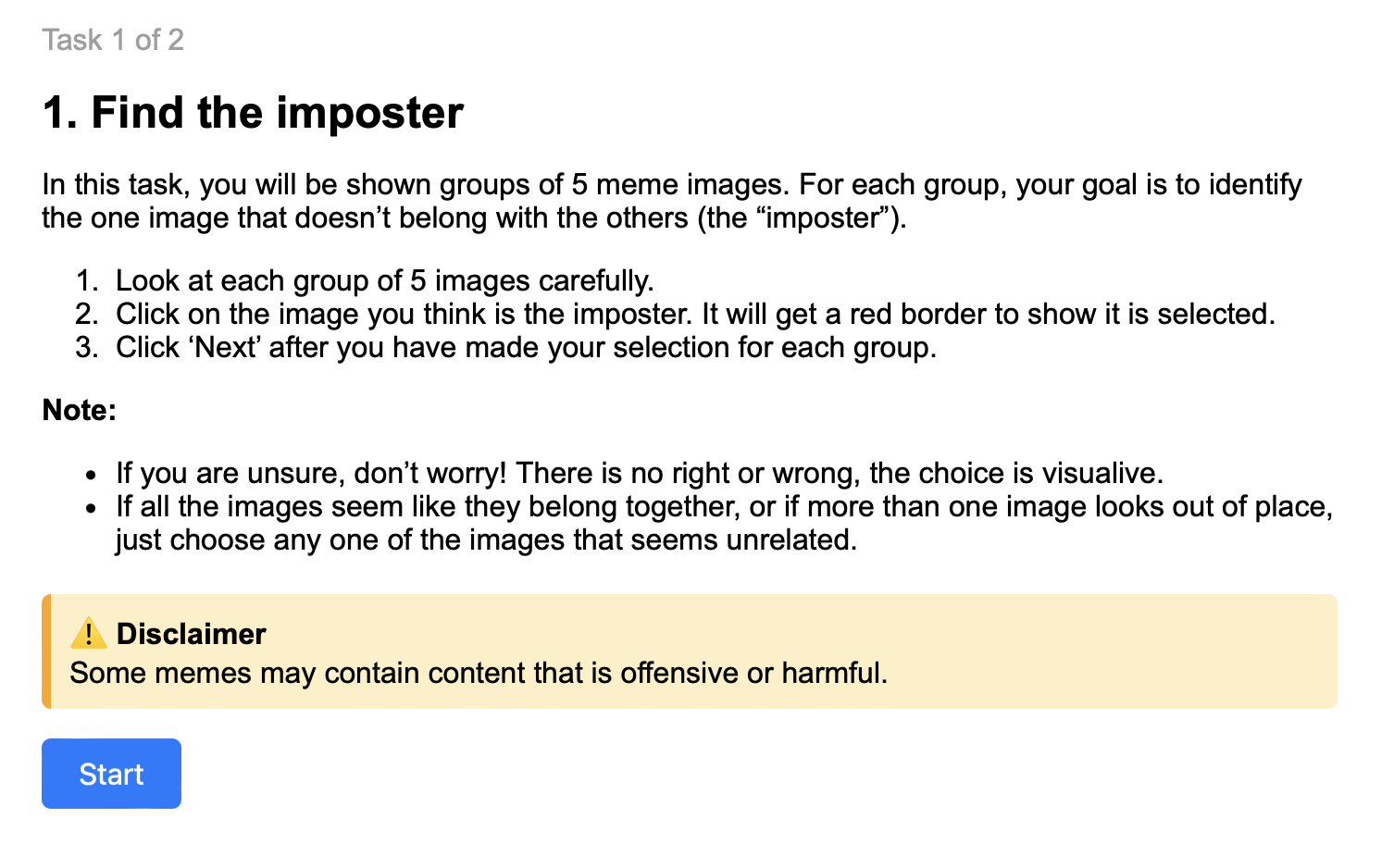}
        \caption{Imposter-Host Task}
        \label{fig:task1_instr}
    \end{subfigure}
    \hfill
    \begin{subfigure}[b]{\linewidth}
        \centering
        \includegraphics[width=\linewidth]{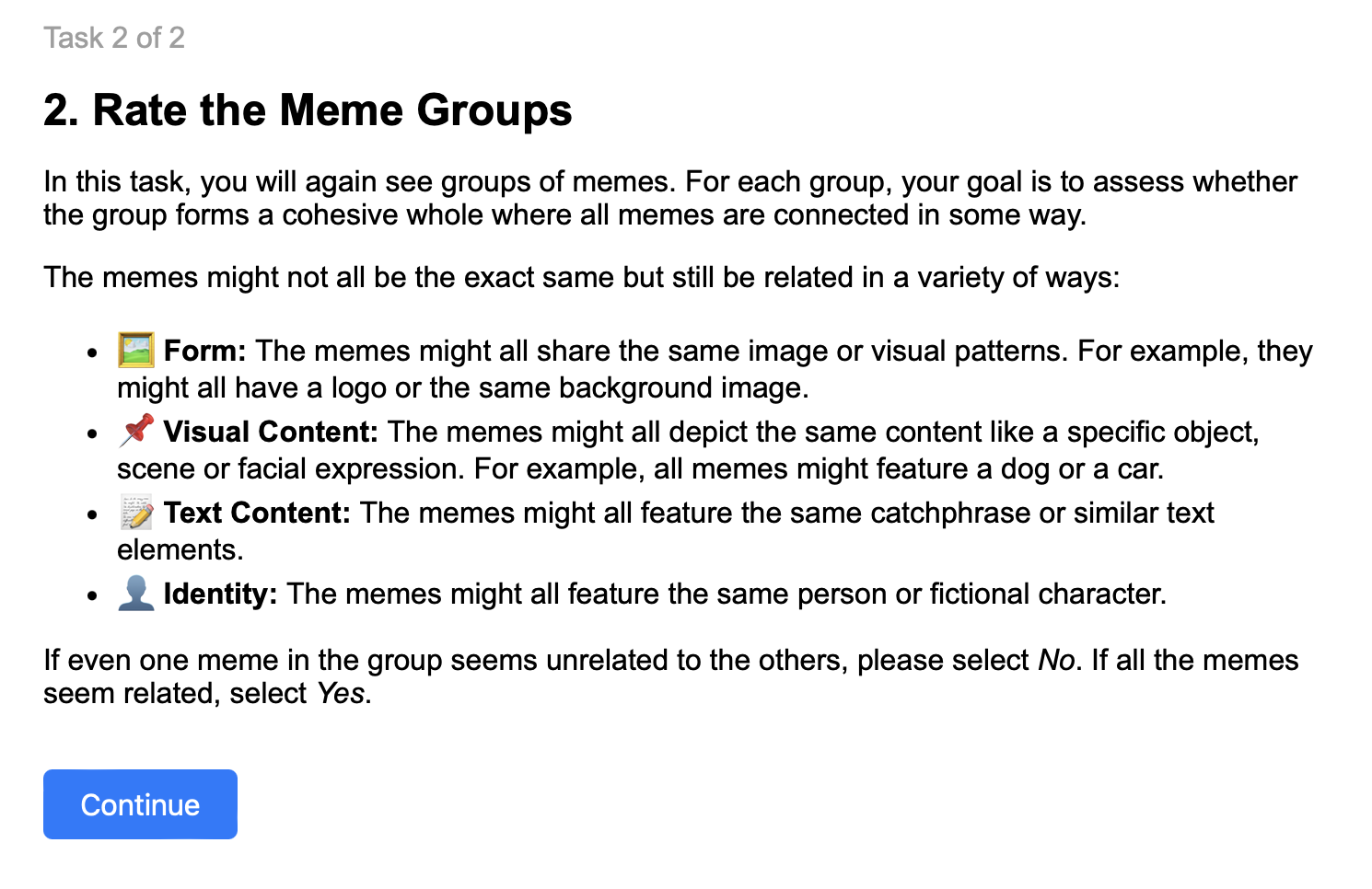}
        \caption{Meme-Cluster Validation Task}
        \label{fig:task2_instr}
    \end{subfigure}
    \caption{Instructions provided to the human judge in each task.}
    \label{fig:instructions}
\end{figure}

\begin{figure*}[!htp]
    \centering
    \includegraphics[width=0.7\linewidth]{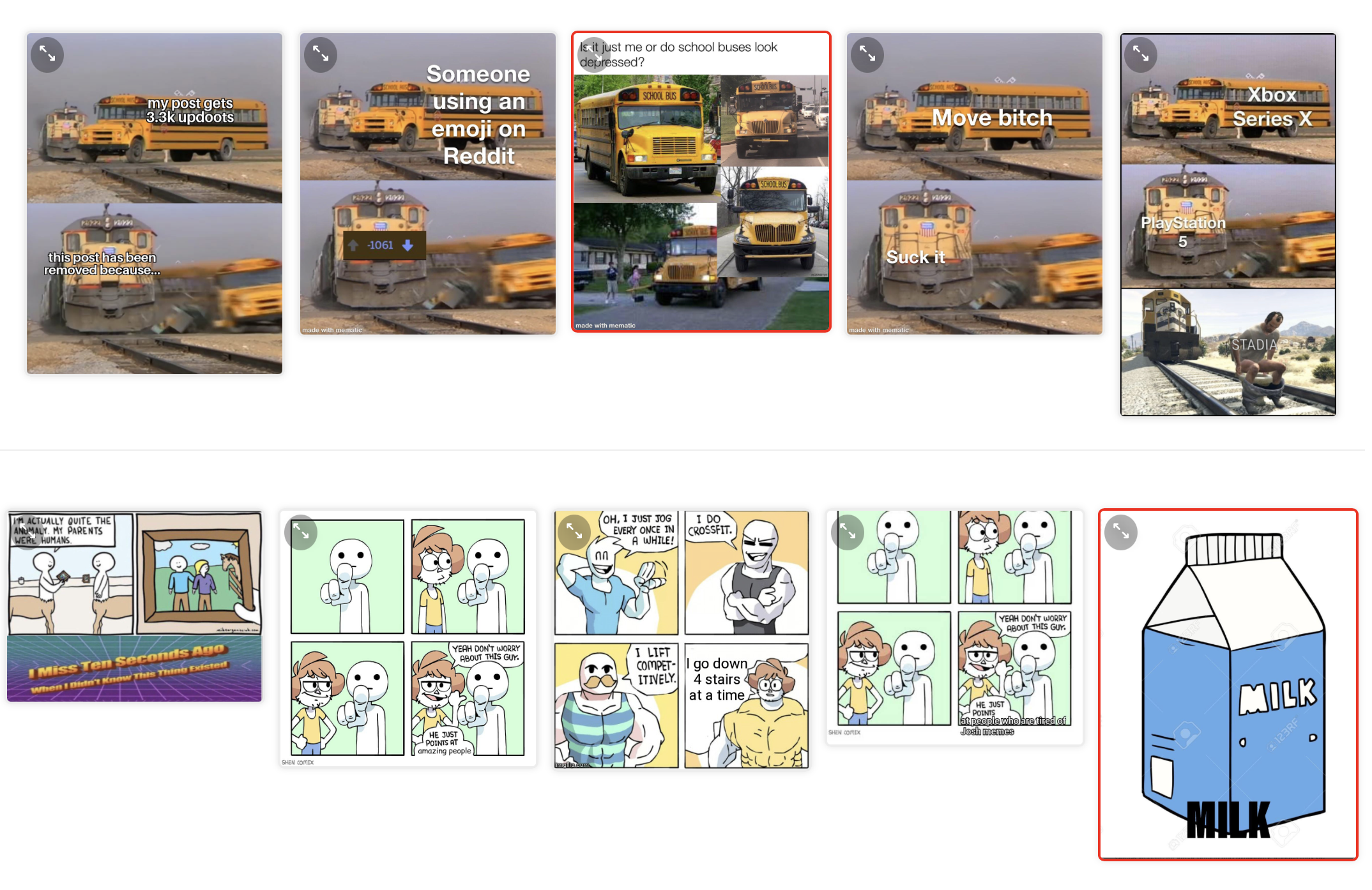}
    \caption{Two examples of the Imposter-Host task interface.}
    \label{fig:imposter-host-setup}
\end{figure*}

\begin{figure*}[!ht]
    \centering
    \includegraphics[width=0.7\linewidth]{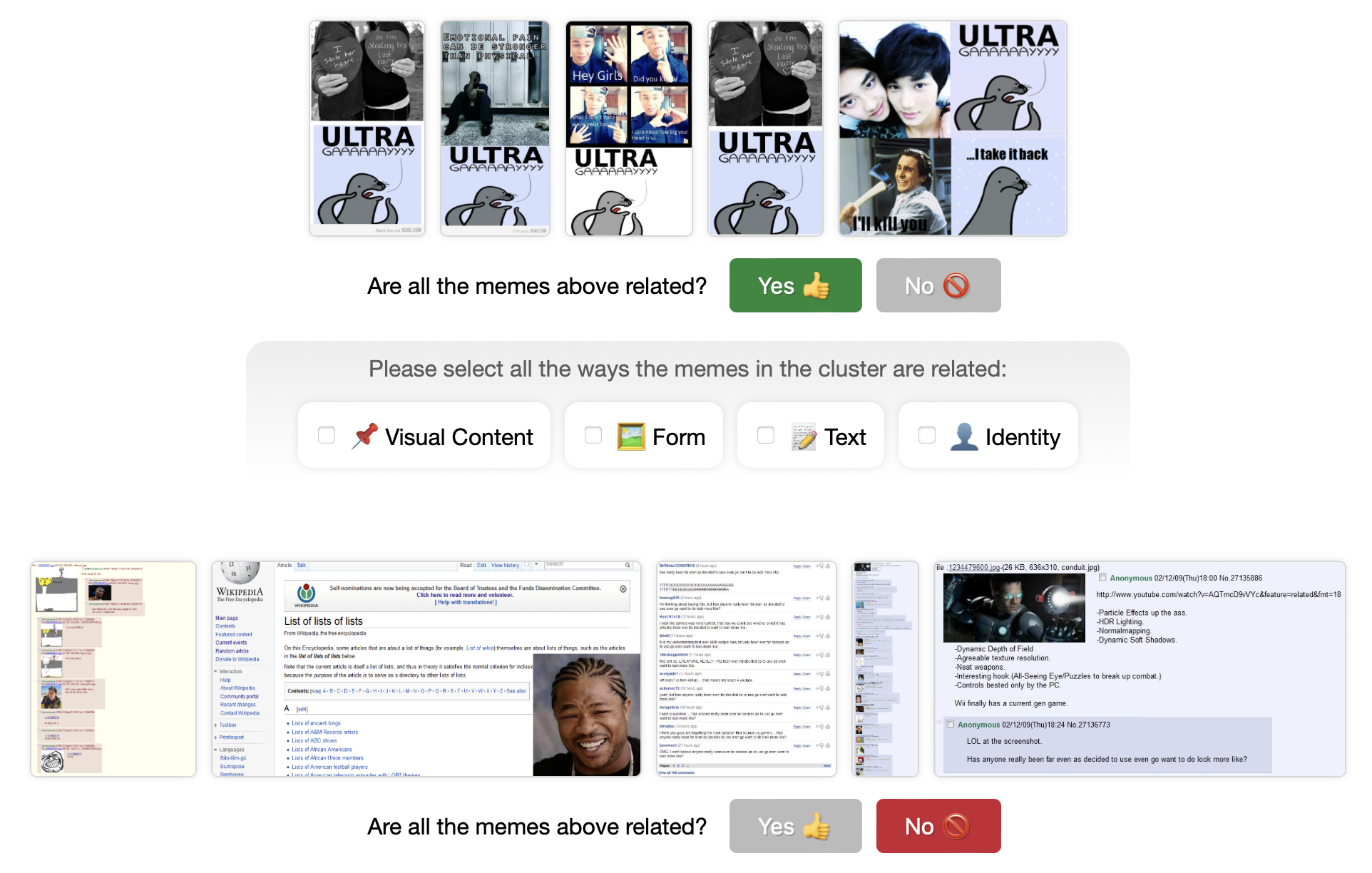}
    \caption{Two examples of the meme cluster validation task.}
    \label{fig:meme-validation-setup}
\end{figure*}

\section{Data Details}

The aggregated human judgments (decoupled from the individual human votes) are stored on a local laptop. Due to their aggregation and \hl{since} no personal information was collected, there is no risk of identifying the participants. As the data is used to validate each approach separately, it cannot be reused \hl{to evaluate} new methods. However, we will make the validation judgments available upon acceptance of our paper. The rest of the data comes directly from KYM and Reddit's Pushshift dump. KYM's license allows for free access and use of its data; however, releasing any adaptations requires written permission from KYM. Meanwhile, Pushshift, is released under the permissive Creative Commons Attribution 4.0 International license.

\subsection{Implementation Details}

For the ViT features, we employ the large variant \textit{google/vit-large-patch32-224-in21k} from HuggingFace.  To extract features from the text embedded in the memes, we first perform Optical Character Recognition through Apple's Vision framework\footnote{https://github.com/straussmaximilian/ocrmac}, chosen for its optimization for our hardware. Features are extracted using the \textit{google-bert/bert-large-uncased} model, also sourced from HuggingFace. For face recognition, we use the Dlib machine learning toolkit's pretrained model \textit{dlib\_face\_recognition\_resnet\_model\_v1.dat}.

SURF is chosen over alternatives like SIFT \cite{lowe.1999} due to its speed and lower memory requirements, with only 64 dimensions per vector.
We extract only the top 1000 keypoint descriptors per image for SURF features to manage the memory requirements. Before extracting SURF features, black boxes are placed over text elements using the EAST text detector \cite{DBLP:journals/corr/ZhouYWWZHL17} to identify regions likely to contain text. The deep learning model generates scores and geometry data adjusted to the original image scale. Black rectangles are drawn over bounding boxes with scores \hl{greater} than 0.5, covering the detected text areas.

While indexing, all vectors are normalized to unit length before being added to the index. We primarily use a simple flat index for high fidelity. However, for SURF local features, we use an Inverted File System with Product Quantization (IVF-PQ) given the large number of vectors. IVF partitions the vector space into 512 Voronoi cells to reduce the search space. \hl{At the same time}, OPQ compresses vectors by decomposing the high-dimensional space into low-dimensional subspaces and quantizing each with 8 bits per subquantizer.

To facilitate an ANN, we build an index for each set of features. The indices are constructed using FAISS (Facebook AI Similarity Search), which enables rapid retrieval of items similar to a query item without \hl{requiring} exhaustive comparison against every item in the dataset \cite{faiss}. 

For constructing adjacency matrices, we update the cells for each image only for the top 100 neighbors in terms of distance. This approach maintains sparsity and aligns with \hl{our dataset's nature, where memes are unlikely} to be meaningfully related to more than 100 other images.

All tools, libraries, and models used in our implementation are publicly available and can be used under permissive licenses. We ran our experiments on a local cluster with A5000 GPUs.





\subsection{\hl{Entropy as a Complementary Metric for Cluster Purity}}
\label{sec:appendix_entropy}

\hl{While the consistency metric reported in the results section evaluates how well clusters align with the dominant ground-truth template (from KYM), it has a limitation: it primarily focuses on the mode of the template distribution within a cluster. It does not fully penalize clusters that exhibit significant mixing across multiple templates, as long as one template remains the most frequent.

We also employ cluster entropy to provide a more nuanced view of cluster quality, particularly for the homogeneity or purity of the generated clusters. Entropy quantifies the uncertainty or impurity in the distribution of ground-truth template labels within each generated cluster. A lower entropy value indicates a purer cluster, dominated by a single template, while a higher entropy value signifies a more mixed cluster with images spread across several different templates.}

\subsubsection*{\hl{Definition}}
\hl{Let \mbox{\( C_t \) be the \( t \)-th} cluster generated by a clustering method. Let \mbox{\( n_{k,t} \)} be the number of images within the cluster \mbox{\( C_t \)} that belong to the \mbox{\( k \)-th} ground-truth template from KYM. The total number of KYM-labeled images in the cluster is \mbox{\( N_t = \sum_{k} n_{k,t} \)}. The proportion of images belonging to the template \mbox{\( k \)} within the cluster \mbox{\( t \)} is given by \mbox{\( p_{k,t} = \frac{n_{k,t}}{N_t} \)}.}

\hl{The entropy \mbox{\( H(C_t) \)} of the cluster \mbox{\( C_t \)} is calculated using the Shannon entropy formula:
\mbox{\(
H(C_t) = -\sum_{k} p_{k,t} \log_2(p_{k,t})
\)}
, where the sum is on all KYM templates \mbox{\( k \)} present in the cluster and, by convention, 
\mbox{\( 0 \log_2 0 = 0 \).}
Minimal entropy \mbox{(\(H=0 \))} occurs when all images in \mbox{\( C_t \)} belong to a single template (perfect purity). 
Maximum entropy occurs when images are uniformly distributed across multiple templates (maximum impurity/heterogeneity). We report the average entropy across all clusters, weighted by cluster size \mbox{(\(N_t\))}, 
similar to the consistency metric.}

\subsubsection*{\hl{Results}}
\hl{Table \ref{tab:entropy_results} presents the average weighted cluster entropy scores for the different methods and feature sets in various numbers of clustered images. Lower scores indicate better cluster purity.}

\begin{table}[!ht]
\centering
\small
\caption{\hl{Average weighted cluster entropy scores (lower is better) across methods and number of images clustered. Evaluated only on images with KYM ground-truth labels.}}
\label{tab:entropy_results}
\begin{tabular}{l l r r r}
\toprule
& & \multicolumn{3}{c}{\hl{\textbf{\# Images clustered}}} \\ \cmidrule(l){3-5}
\textbf{\hl{Clustering method}} & \textbf{\hl{Feature set}} & \hl{\textbf{5000}} & \hl{\textbf{8500}} & \hl{\textbf{11000}} \\ \midrule

\multirow{4}{*}{\hl{Standard}} & \hl{ViT}    & \hl{1.75} & \hl{3.19} & \hl{4.06} \\
& \hl{Global}   & \hl{0.96} & \hl{2.50} & \hl{3.35} \\
& \hl{Local}    & \hl{0.91} & \hl{1.72} & \hl{2.85} \\
& \hl{Combined} & \hl{\textbf{0.29}} & \hl{1.90} & \hl{2.81} \\
\midrule
\multirow{4}{*}{\hl{Template-based (ours)}} & \hl{ViT} & \hl{1.75} & \hl{1.12} & \hl{0.91} \\
& \hl{Global}  & \hl{0.96} & \hl{0.56} & \hl{0.45} \\
& \hl{Local}   & \hl{0.91} & \hl{0.55} & \hl{0.45} \\
& \hl{Combined} & \hl{\textbf{0.29}} & \hl{\textbf{0.24}} & \hl{\textbf{0.21}} \\

\bottomrule
\end{tabular}
\end{table}

\subsubsection*{\hl{Discussion}}
\hl{The entropy results largely corroborate the findings observed with the consistency metric. Our template-based clustering approach consistently yields a lower average entropy (Table \ref{tab:entropy_results}), indicating purer clusters with fewer mixing between different ground-truth templates than standard clustering. This advantage becomes more pronounced as more images are clustered. Whereas the standard clustering entropy increases sharply, indicating an increase in impurity, our method maintains or even improves the purity of the cluster.}

\hl{Furthermore, consistent with prior results, the `Combined` feature set used with our template-based method achieves the lowest entropy, demonstrating that integrating multiple similarity dimensions leads to the most homogeneous and semantically well-defined clusters. In essence, entropy analysis reinforces that our proposed methodology produces clusters dominated by the correct template (as shown by consistency) and significantly less contaminated by unrelated templates.}

\subsection{Additional Results}

The number of clusters discovered by our method using various feature sets is shown in Table \ref{tab:num_clusters}.

\begin{table}[!ht]
\centering
\small
\begin{tabular}{l r r r}
\toprule
& \textbf{5000} & \textbf{8500} & \textbf{11000} \\ \midrule
ViT & 832 & 932 & 816  \\
Global  & 933 & 1,071 & 981 \\
Local   & 865 & 974 & 727    \\
Combined & 1,003 & 1,144 & 1,031 \\

\bottomrule
\end{tabular}
\caption{Number of clusters discovered with our method using various feature sets for 5000, 8500, and 11000 memes.}
\label{tab:num_clusters}
\end{table}

\hl{The} results with DBSCAN as a clustering algorithm instead of Louvain clustering are shown in Table \ref{tab:dbscan}.


\begin{table}[!t]
\centering
\small
\begin{tabular}{l l r r r}
\toprule
& & \multicolumn{3}{c}{\textbf{\# Images clustered}} \\ \cmidrule(l){3-5} 
\textbf{Clustering method} & \textbf{Feature set} & \textbf{5000} & \textbf{8500} & \textbf{11000} \\ \midrule

\multirow{4}{*}{Standard} & ViT    & 0.68  &  0.29  &  0.10   \\
& Global   & 0.83  &  0.45  &  0.13              \\
& Local    & 0.70 &   0.36 &   0.17              \\ 
& Combined & \textbf{0.94}               & 0.47   & 0.15               \\
\midrule
\multirow{4}{*}{Template-based (ours)} & ViT & 0.68  &  0.68  &  0.65 \\
& Global  & 0.83  &  0.81  &  0.78              \\
& Local   & 0.70 &   0.84   & 0.79        \\
& Combined & \textbf{0.94}  &  \textbf{0.89}  &  \textbf{0.87}     \\

\bottomrule
\end{tabular}
\caption{Consistency scores across methods and \# images clustered when using DBSCAN clustering algorithm. 
The best results are shown in bold. The numbers of clusters for various feature sets are given in the Appendix.}
\label{tab:dbscan}
\end{table}

\subsection{Practical Application: Dynamic Meme Retrieval per Similarity Dimensions}

\hl{We developed a user interface to} demonstrate how our method facilitates dynamically serving the most relevant memes to a query. Our methodology proceeds as follows.

\begin{enumerate}
    \item \textbf{Template Indexing}: All identified templates are enriched with descriptive information. We detect web entities \hl{for each template} and combine this with the images as input for a multimodal large language model (LLM). The model is prompted to generate short descriptions and keywords (refer \hl{to} Listing \ref{lst:prompt} for the prompt). The captions are subsequently embedded using another language model and indexed using FAISS. 
    \item \textbf{Template Similarity Search:} For a given query, we embed it using the same text embedding model as the templates. The templates \hl{most similar} to the query are identified through the FAISS nearest-neighbor search.
    \item \textbf{Image-Template Matching:} Using the identified templates, we calculate their similarity vectors using the appropriate adjacency matrix. Images with the highest \hl{sum of} similarity to the templates are retrieved. For instance, if the query pertains to a specific person or fictional character, the adjacency matrix belonging to the `identity' similarity dimension is employed to find non-templatic memes with the same person.
\end{enumerate}

This method enables the dynamic delivery of contextually relevant memes. For example, querying `Joe Biden' yields a template to identify similar memes based on the `identity' feature set, while querying `car' uses the `visual content' feature set, as it \hl{does not} pertain to individuals or specific visual elements. \hl{Selection of dimensions may be done through NLP methods such as Entity Type Classification and Named Entity Recognition, through a large language model with function-calling or other automated techniques. Currently, we have the user manually select these dimensions.}
Figure \ref{fig:examples_demo} illustrates this with an example, and Figure \ref{fig:indexed_tmplate} shows an indexed meme template and its LLM-generated description. Prompt-based methods may also extract additional features, such as stance, for better clustering. Furthermore, enriched information enables the integration of background knowledge in the clustering process. For example, memes that feature various fictional characters from the same television show may be linked in this manner. However, the effectiveness of these methods requires further investigation.


\FloatBarrier
\begin{lstlisting}[caption={Prompt for enriching templates.}, label={lst:prompt}]
You are an expert in internet memes. You are given a number of examples images of a meme. While the images may differ individually, they should share a common meme template, e.g. share patterns and elements composing the ideas and behaviors in memes. Your task is to identify the meme template, and describe it.

Give a suggested title for the meme template, a description of the meme, and identify the entities and tags that are relevant to the meme. Here are three examples:

{
    "title": "Pajama Kid",
    "description" :  "A photoshop featuring a yearbook photograph of a young boy wearing SpongeBob Squarepants-themed pajamas with a resigned expression on his face.",
    "entities": [
        {
            "entity": "boy",
            "type": "person"
        }
    ],
    "tags": [
        "boy",
        "kid",
        "pajamas",
        "SpongeBob Squarepants",
        "resigned expression",
    ]
}

{
    "title": "Drake The Type Of Guy",
    "description" :  "Fan-written factoids that are presented as the personality traits of the rapper Drake. Pokes fun at the rapper's stage persona as being emotionally sensitive and even effeminate, which goes against the alpha male stereotype that is still prevalent in hip-hop.",
    "entities": [
        {
            "entity": "Drake",
            "type": "person"
        }
    ],
    "tags": [
        "Drake",
        "rapper",
        "emotional",
        "sensitive",
        "effeminate"
    ]
}

{
    "title": "King Charles' Portrait",
    "description" :  "Portrait of King Charles III, done by artist Jonathan Yeo, features Charles' face while his body blends into a bright red background. It appears demonic, as though Charles was in the fires of hell.",
    "entities": [
        {
            "entity": "King Charles III",
            "type": "person"
        },
        {
            "entity": "Jonathan Yeo",
            "type": "person"
        },
        {
            "entity": "portrait painting",
            "type": "object"
        }
    ],
    "tags": [
        "King Charles III",
        "Jonathan Yeo",
        "portrait painting",
        "demonic",
        "hell"
    ]
}

Always use this JSON format.

Note: only incorporate the entities and tags that are relevant to all the images in the meme template, not just one image.

You are given the following images:
[images]

\end{lstlisting}

\begin{figure}[!ht]
    \begin{subfigure}[b]{\linewidth}
        \centering
        \includegraphics[width=\textwidth]{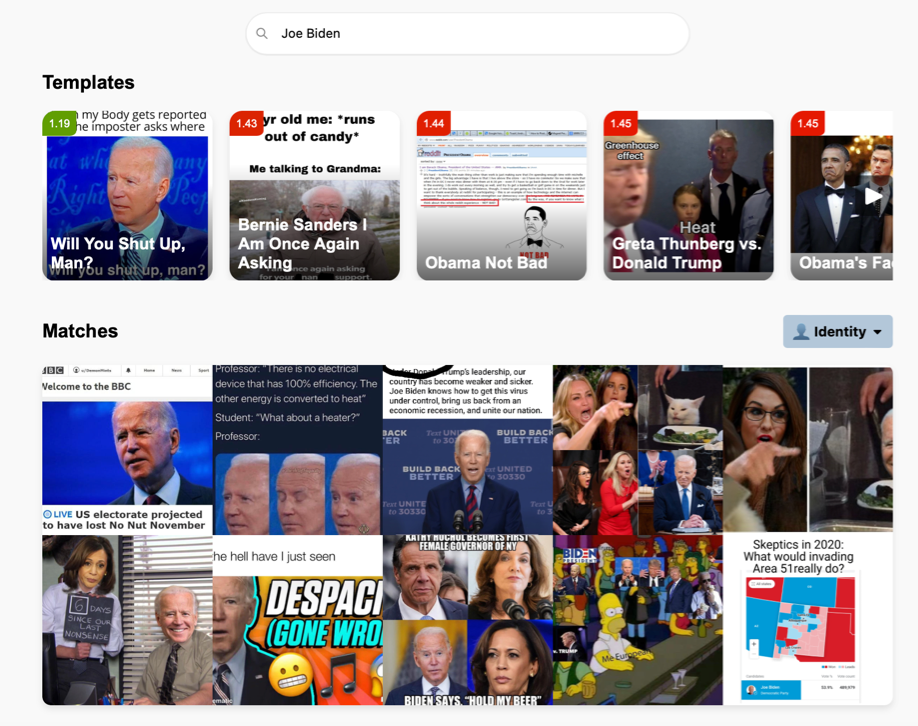}
        \caption{}
        \label{fig:biden}
    \end{subfigure}
    \hfill
    \begin{subfigure}[b]{\linewidth}
        \centering
        \includegraphics[width=\textwidth]{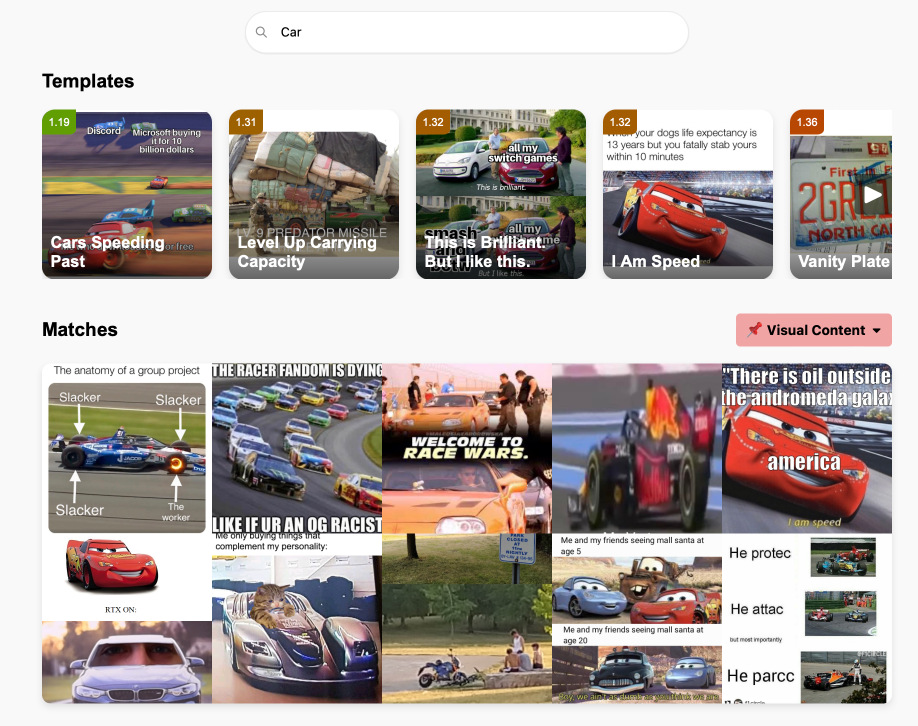}
        \caption{}
        \label{fig:car}
    \end{subfigure}
    \caption{Templates and memes found for queries}
    \label{fig:examples_demo}
\end{figure}

\begin{figure}
    \centering
    \includegraphics[width=\linewidth]{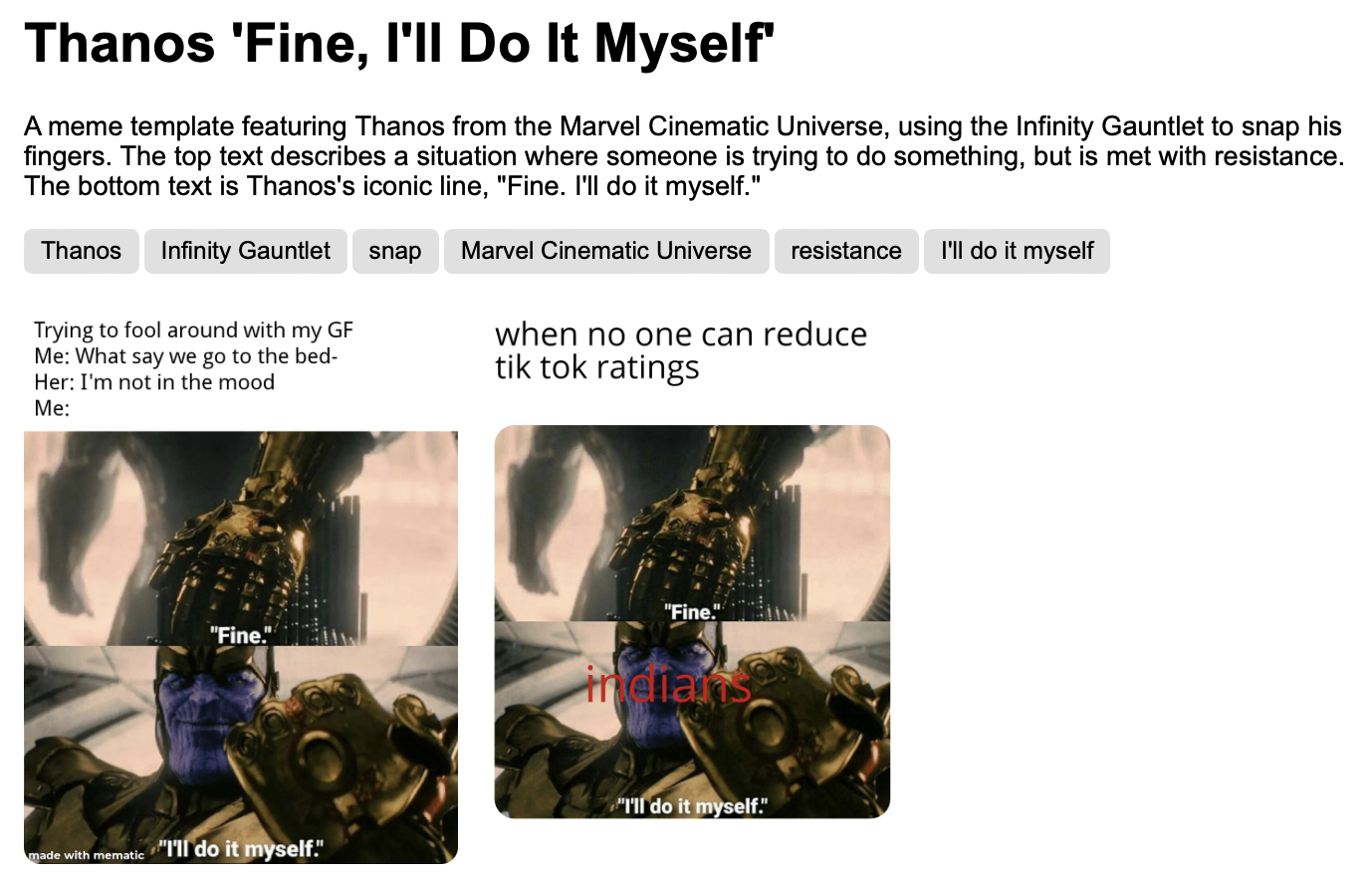}
    \caption{Example of an indexed template after information enrichment.}
    \label{fig:indexed_tmplate}
\end{figure}

\end{document}